\documentclass{article}

\usepackage{microtype}
\usepackage{graphicx}
\usepackage{subcaption}
\usepackage{booktabs} 

\usepackage{hyperref}

\usepackage[preprint]{icml2026}

\usepackage{amsmath}
\usepackage{amssymb}
\usepackage{mathtools}
\usepackage{amsthm}

\usepackage[capitalize,noabbrev]{cleveref}

\theoremstyle{plain}
\newtheorem{theorem}{Theorem}[section]

\newtheorem{lemma}[theorem]{Lemma}
\newtheorem{corollary}[theorem]{Corollary}
\theoremstyle{definition}
\newtheorem{definition}[theorem]{Definition}
\newtheorem{assumption}[theorem]{Assumption}
\theoremstyle{remark}
\newtheorem{remark}[theorem]{Remark}

\usepackage[textsize=tiny]{todonotes}

\usepackage{enumitem}
\usepackage{algorithm}
\usepackage{algorithmic}
\usepackage{adjustbox}
\usepackage{amsmath, amssymb, amsthm}  
\usepackage{bm}                         
\usepackage{mathrsfs}                   

\usepackage{booktabs}                   
\usepackage{threeparttable}             
\usepackage{multirow}                   
\usepackage{caption}                    
\usepackage{subcaption}                 

\usepackage{graphicx}                   
\usepackage{adjustbox}                  
\usepackage{tikz}                       

\usepackage{natbib}                     

\usepackage{xcolor}                     
\usepackage{hyperref}                   

\usepackage{pifont}
\newcommand{\cmark}{\ding{51}}  
\newcommand{\xmark}{\ding{55}}  
\usepackage{colortbl}
\definecolor{lightpink}{rgb}{1,0.9,0.9}
\definecolor{lightblue}{rgb}{0.8,0.9,1.0}
\usepackage[english]{babel} 
\usepackage{csquotes}
 
\begin{document}

\twocolumn[
  \icmltitle{Fine-Tuning Flow Matching via Maximum Likelihood Estimation of Reconstructions}



  \icmlsetsymbol{equal}{*}

\begin{icmlauthorlist}
	\icmlauthor{Zhaoyi Li}{yyy,comp,sch}
	\icmlauthor{Jingtao Ding}{tsinghua}
	\icmlauthor{Yong Li}{comp,tsinghua}
	\icmlauthor{Shihua Li}{yyy,sch}
\end{icmlauthorlist}

\icmlaffiliation{yyy}{School of Automation, Southeast University}
\icmlaffiliation{sch}{Key Laboratory of MCCSE of the Ministry of Education, Southeast University}
\icmlaffiliation{comp}{Beijing Zhongguancun Academy}
\icmlaffiliation{tsinghua}{Department of Electronic Engineering, Tsinghua University}

\icmlcorrespondingauthor{Shihua Li}{lsh@seu.edu.cn}
%
%

  \icmlkeywords{Machine Learning, ICML}

  \vskip 0.3in
]



\printAffiliationsAndNotice{}  

\begin{abstract}

Flow Matching (FM) models achieve remarkable results in generative tasks. Building upon diffusion models, FM's simulation-free training paradigm enables simplicity and efficiency but introduces a train-inference gap: model outputs cannot be assessed during training. 
Moreover, the straight flow assumption suffers from some inherent limitations.
To address this, we propose to fine-tune FM via Maximum Likelihood Estimation (MLE) of reconstructions---enabled by FM's smooth ODE formulation, unlike the stochastic differential equations (SDEs) in diffusion models.
We first theoretically analyze the relationship between training loss and inference error in FM under numerical precision constraints. We then propose an easy-to-implement fine-tuning framework based on MLE of reconstructions, with flexibility for sophisticated extensions. Building on this, we incorporate a generalized artificial viscosity term that enhances flow stability and robustness, accompanied by a direct parameterization method and rigorous theoretical guarantees.
Experiments demonstrate our method's effectiveness across diverse settings: a toy example provides mechanistic insights into the fine-tuning process, while large-scale evaluations on meteorological forecasting and robotic manipulation policies validate reliable performance improvements.
\end{abstract}

\section{Introduction}



Deep generative models refer to a category of deep learning techniques designed to approximate and generate samples from an unknown underlying data distribution. A mainstream paradigm is to learn a mapping between a fixed (e.g., standard normal) distribution and the data distribution.
This category notably includes diffusion models, a leading approach for many generative tasks across vision \cite{rombach2022high,ho2022video}, policy \cite{chi2023diffusion}, and spatio-temporal data \cite{li2022generative,yuan2023spatio}. 
The mathematical principles behind diffusion can be described by SDEs \citep{songscore}. Naturally, we can also establish the relationship between noise and samples through ODE trajectories to simplify the model. 
This inspired the development of the Flow Matching (FM) algorithm \citep{lipman_flow_2022, liuflow, albergo2023building}. 
FM has garnered extensive attention \cite{irvin2025spatiotemporal,rohbeck2025modeling,wang2025vaflow,esser2024scaling,hamad2025flow}, particularly emerging as a mainstream approach in robot policy due to its fast inference speed \citep{black2410pi0, zhang2025flowpolicy, braun2024riemannian,
	chisarilearning, zhang2024affordance}. 

FM, which inherits the characteristics of diffusion, employs a simulation-free training approach.
This means that during the training phase, we only train some intermediate variables, e.g., vector filed \citep{lipman_flow_2022}, score \citep{song2019generative}, and noise (or the previous state) \citep{ho2020denoising}. We cannot directly observe and optimize the final output from these difference or differential terms.
In contrast, other generative models directly includes generated samples in their training loss. In Variational Autoencoder (VAE), we contain the rescontruction error \citep{kingma2013auto}. 
In normalizing flow, we use the Maximum Likelihood Estimation (MLE) of the final output generated by the model \citep{rezende2015variational, chen_neural_2018}.
Generative Adversarial Networks (GANs) are similar except they use adversarial training to replace the likelihood function \citep{goodfellow2014generative}.

\begin{figure}[tb]
	\centering
	\begin{adjustbox}{scale=0.8}
		\begin{subfigure}[b]{0.6\linewidth} 
			\centering
			\includegraphics[angle=-90, width=\linewidth]{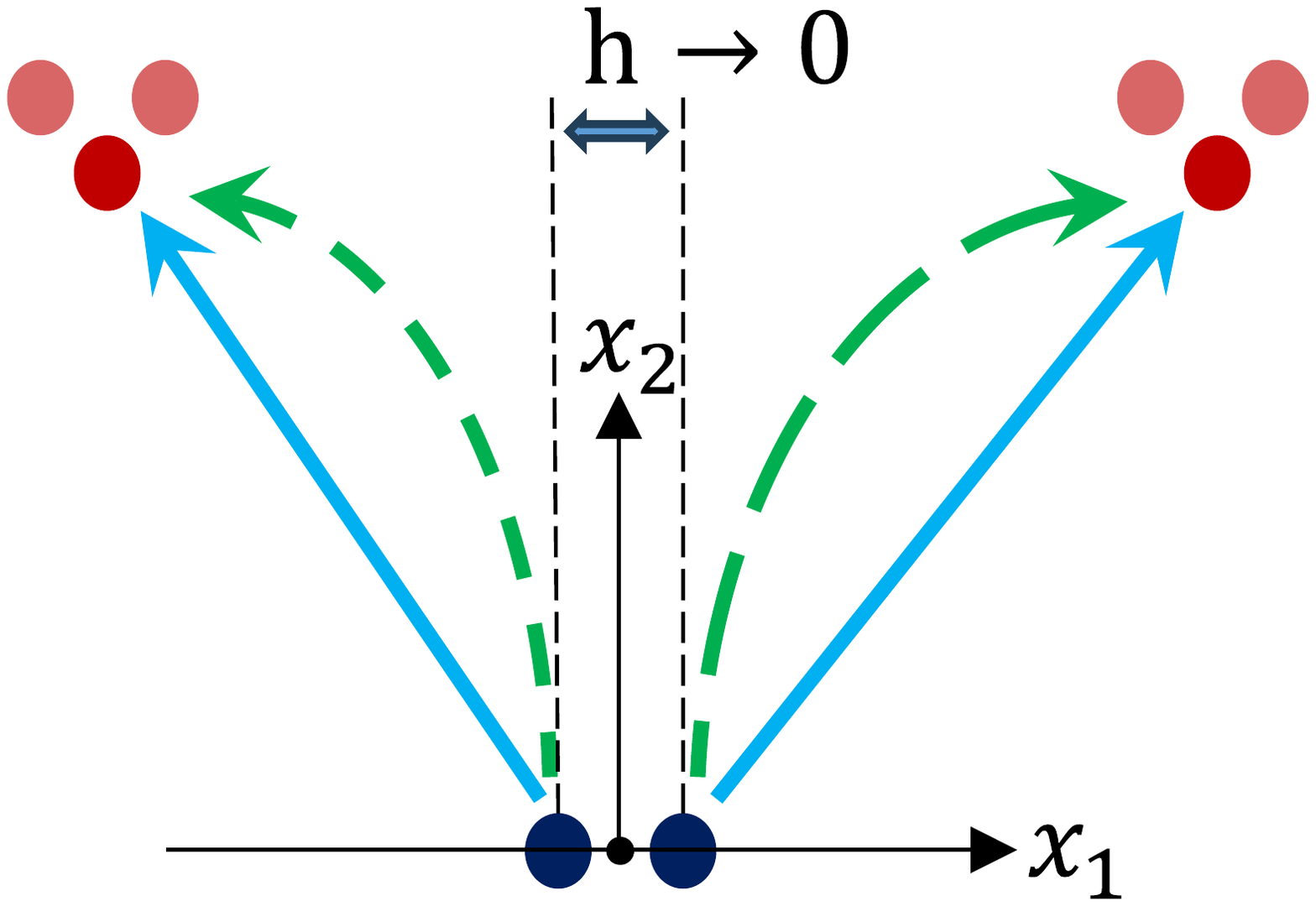}
			\caption{Straight Flow}
			\label{fig:flow_dis}
		\end{subfigure}		
		\begin{subfigure}[b]{0.34\linewidth} 
			\centering
			\includegraphics[angle=0, width=\linewidth, trim=0 -0.7cm 0 0, clip]{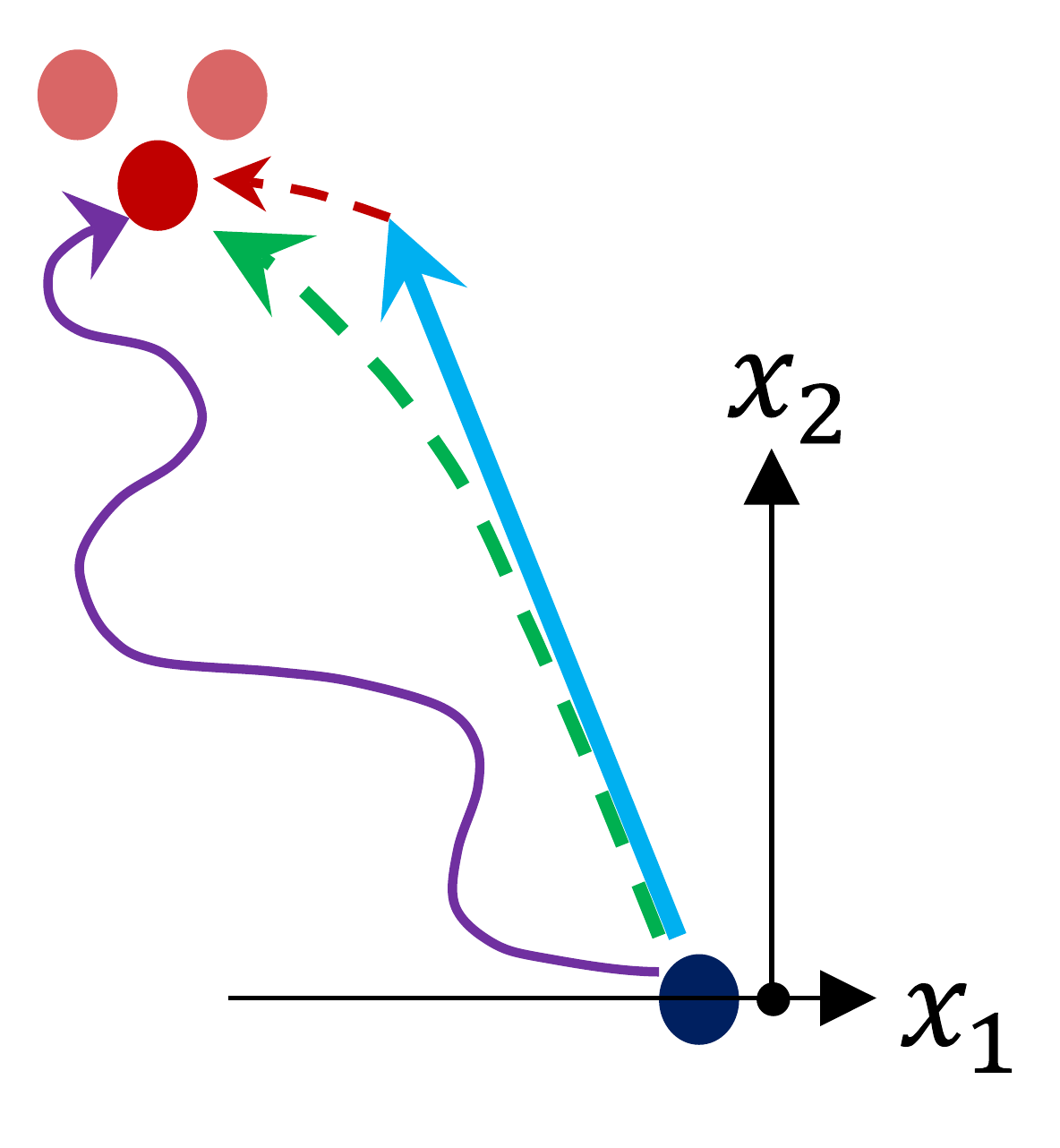}
			\caption{Finetuned Flow}
			\label{fig:flow_f}
		\end{subfigure}		
	\end{adjustbox}
    \\
	\begin{adjustbox}{scale=0.8}
		\begin{subfigure}[b]{0.6\linewidth}
			\centering
			\includegraphics[angle=-90, width=\linewidth, trim=0 -1cm 0 0, clip]{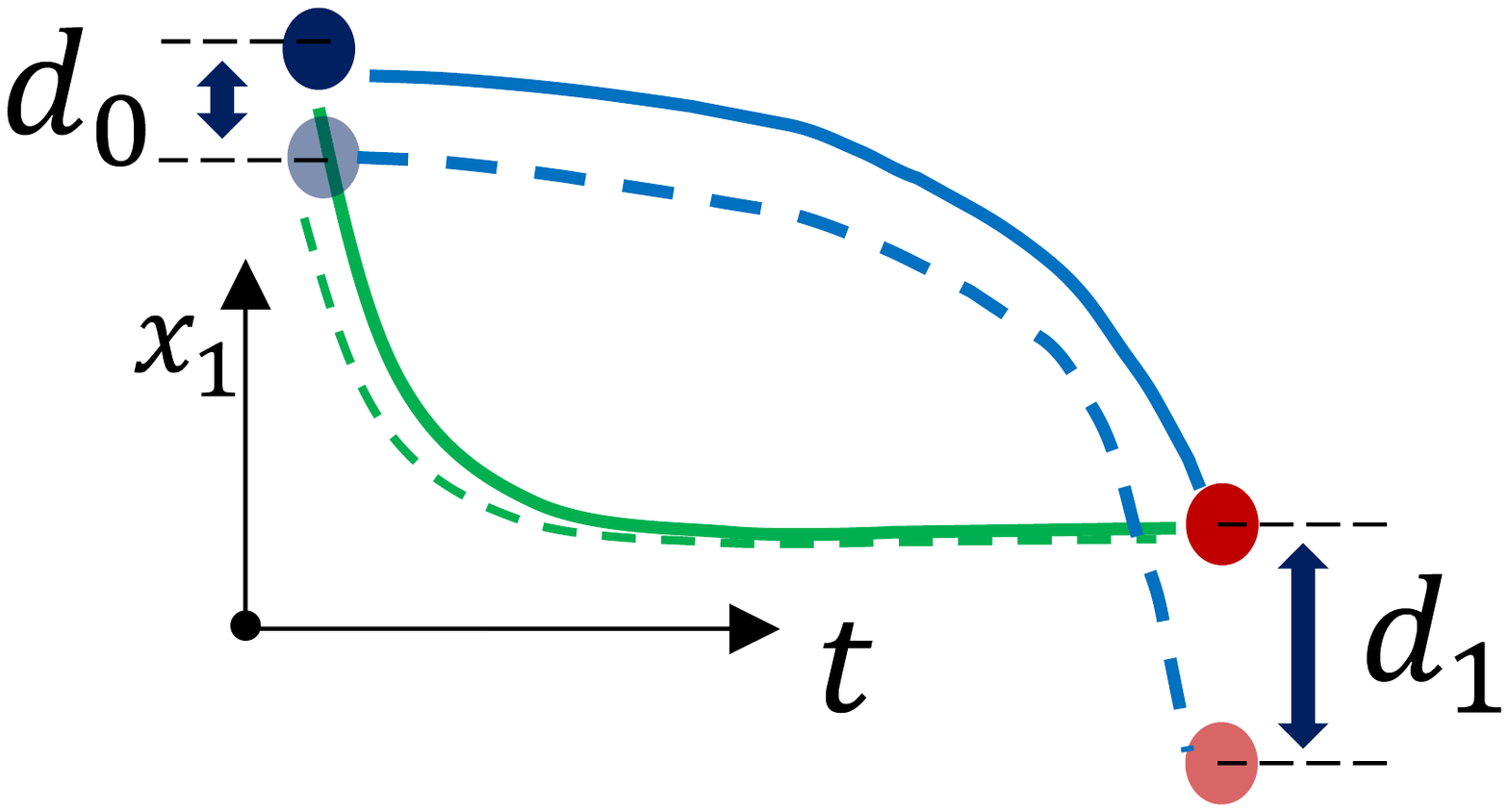}
			\caption{1D Contraction}
			\label{fig:flow_c}
		\end{subfigure}
		\begin{subfigure}[b]{0.4\linewidth}
			\centering
			\includegraphics[angle=-90, width=\linewidth]{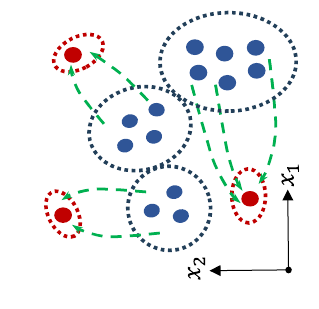}
			\caption{2D Contraction}
			\label{fig:flow_con_2dim}
		\end{subfigure}
	\end{adjustbox}
	\caption{Fine-Tuning Your Flow: A Visual Explanation.
		\textbf{(a)} illustrates that over-pursuing straightness (blue lines) leads to discontinuities in the vector field, i.e., $f(0^+,0)\neq f(0^-,0) $. 
        By comparison, the green line depicts a more stable flow. 		
		\textbf{(b)} plots the pre-trained FM path (blue line), the fine-tuned flow (green line), the flow fine-tuned with residuals (red line), and the flow trained entirely with MLE (purple line). 
		An oversimplified assumption like "the straight path" can lead to underfitting (blue line).
		A fine-tuned model converges to a local optimum near the pre-trained model, thereby improving its fit to the sample while maintaining path simplicity (green line). 
		By contrast, the CNF (purple line) often fits the samples with overly complex trajectories.
		The red line utilizes a residual network to learn the remaining residual components.
		\textbf{(c)} plots the variation curve of component $x_1$ over time. Here, compared with the blue line, the green one represents a ``contracting" trajectory. When subjected to a minor disturbance $d_0$ 
        , a contracting trajectory still tends to stabilize around a similar solution, showing superior stable performance. 
		\textbf{(d)} illustrates contraction in 2-dimensional space. (Blue) points in different contraction regions will converge to different (red) destinations. Points in the same contraction region behave stably and robustly.
	}
	\label{fig:flow_three}
\end{figure}

Therefore, though offering relatively fast training speed, we have no knowledge of how the real samples are actually generated during the simulation-free training phase. 
This implies the existence of a gap between its training and inference  phases.
Such a gap can considerably impact scenarios that require high precision, such as robotic manipulation or spatio-temporal numerical tasks.
By comparison, the Action Chunking with Transformers (ACT) architecture \citep{zhao2023learning}, built upon a VAE with integrated reconstruction error, performs remarkably well in fine manipulation tasks.
Moreover, we point out that FM's over-pursuit of straight paths may render the system stiff or even lead to discontinuous vector fields (see Figure \ref{fig:flow_dis}), resulting in numerical instability and model unreliability. Although in practice, the stochastic noise introduced by batching and early stopping technique can smooth the model output and mitigate this phenomenon, it comes at the cost of underfitting.
Fortunately, due to the smoothness of ODE trajectories \citep{chen_neural_2018}, it is feasible to fine-tune FM directly by reconstruction error. There are multiple ways to track parameter gradients \citep[Chapter 5]{kidger2021neural} such as adjoint sensitivity method.
In contrast, the SDE-based diffusion models can only estimate—not directly optimize—the maximum likelihood \cite{song2021maximum}.


The schematic diagrams in Figure \ref{fig:flow_three} outline the core principles of our proposed methodology.
Fine-tuning enhances the representational capacity of vector fields while preserving their simplicity. Notably, owing to the flexibility of the MLE fine-tuning framework, we can incorporate additional designs to induce models with favorable properties. 
Inspired by Scientific Computing, we introduce a generalized viscosity during fine-tuning, which endows the flow with enhanced stability and robustness against minor perturbations (i.e., contraction, a concept from control theory).
This paper is organized as follows. 
In Section \ref{sec:motivation} we theoretically analysis the relation between training loss and inferring error . 
Then in Section \ref{sec:method}, we propose a framework of fine-tuning FM via MLE of reconstruction error, and introduce an generalized viscosity for robustness.
Section \ref{experiments} shows the experimental results.

Our primary contributions are: 1) the first theoretical analysis of training-inference error in FM under numerical accuracy constraints; 2) an easy-to-implement fine-tuning framework centered on a concrete MLE loss, and a generalized artificial viscosity term within this framework to enhance robustness; 4) rigorous theoretical proofs for all proposed methods; and 5) comprehensive experiments from toy examples to large-scale meteorological and robotics tasks.

\section{Related work}

\paragraph{MLE Training of ODE Model.}
Neural Ordinary Differential Equations (NODE), or Continuous Normalizing Flow (CNF) in the context of generation tasks \citep{chen_neural_2018}, also adopt the ODE model. 
In contrast to FM's dependence on predefined reference paths, these methods utilize an optimization framework grounded in MLE, using the change of variables theorem with a tractable noise distribution to compute probabilities.
Consequently, they tend to generate complex and intractable vector fields. Significant efforts have been devoted to its improvement \cite{dupont2019augmented, finlay2020train}. 
However, these methods still fail to match the efficiency of simulation-free approaches like FM. 


\paragraph{Optimal-Transport Conditional FM.}
The original FM method constructs paths based on individual sample pairs, which may lead to twisted, entangled, and complex trajectories across the overall distribution. Several studies \cite{liu_rectified_2022,tong2024improving,pooladian2023multisample} have been dedicated to addressing this issue. 
The latter two consider the joint probability between noise and samples within a batch and approximate the optimal path between global distributions. 
Given the complexity of MLE training, this step is important for enhancing the performance of our method and accelerating its convergence.

\paragraph{Stability of ODE Model.} 
Current investigations regarding the stability of Neural ODEs are largely confined to the domain of classification \cite{llorente2021deep,mei2024controlsynth,kang2021stable,massaroli2020stable}, while Mei et al. \yrcite{mei2024controlsynth} experimentally explore the ability of their network to model dynamics.
Some works also address stability analysis in dynamic modeling, such as scenarios involving prior physical constraints \citep{white2023stabilized} or external inputs \cite{li2025icode}.
Our work primarily focuses on the generative domain, encompassing prediction or decision-making tasks,
which also represent the mainstream approaches for these tasks nowadays.

\paragraph{Fast Inference of FM.}
Another popular research direction aims at improving inference speeds, achieving comparable inference quality in single or few steps to that of multi-step inference.
Some methods rely on distillation \cite{geng2023one,salimans2022progressive,sauer2024adversarial,luo2023diff,yin2024one,zhou2024score}
which inevitably incurs the cost of distillation error. 
Other works impose constraints directly on the vector field \cite{song2023consistency,song2023improved,gengconsistency,lusimplifying,yang2024consistency}. However, from the perspective of dynamical systems, such methods tend to increase system stiffness or even discontinuities. Our work amis to enhance the upper bound of the model's inference capability by constructing a better-posed vector field.



\section{Motivations of Fine-Tuning FM}\label{sec:motivation}

\subsection{Preliminaries and Notation }
Let $\mathbb{R}^d$ denote the data space with data points $x = (x^1, \ldots, x^d) \in \mathbb{R}^d$. 
Denote the probability path $p_t : \mathbb{R}^d \rightarrow \mathbb{R}_{+}$, which is a time dependent (for $t \in [0,1]$) probability density function, i.e., $\int p_t(x) dx = 1$, and a time-dependent vector field, $u_t : [0,1] \times \mathbb{R}^d \rightarrow \mathbb{R}^d$. A vector field $u_t$ constructs a time-dependent diffeomorphic map, called a flow, $\psi : [0,1] \times \mathbb{R}^d \rightarrow \mathbb{R}^d$, defined via the ordinary differential equation (ODE):
\begin{equation}\label{eq:FM}
	\frac{d}{dt} \psi_t(x(0)) = u_t(\psi_t(x(0)),\bm o), \quad \psi_0(x(0)) = x_0.
\end{equation}
Here $\bm o$ is the outside information or condition, and we may omit it when there is no ambiguity.
Given two marginal distributions $q_0(x_0)$ and $q_1(x_1)$ for which we
would like to learn a model to transport between, FM seeks to optimize the simple regression objective $\mathbb{E}_{t, p_t(x)} \left\| v_t(x; \theta) - u_t(x) \right\|^2$, where $u_t(x)$ is the is a vector field that generates a probability path $p_t$ under the two marginal constraints, $v_t(x; \theta)$ is the parametric vector field. 
Let $\phi_t(x_0)$ be the solution to the ODE $\frac{d}{dt} \phi_t = v_t(\phi_t; \theta)$ with initial value $\phi_0 = x_0$.
To obtain the numerical solution, let us define $N$ as the total number of discrete steps, $t_i$ as the $i$-th time point, $\tau_i=t_{i+1}-t_i$ as $i$-th interval. Different time points typically involve different step sizes if we employ adaptive step size algorithms, as is often the case. 
We use $\hat{\phi}_n(x_0)$ to represent the corresponding numerical solution at time $t_n$, with the model error $\varepsilon_n(x_0):=\psi_{t_n}(x_0)-\hat{\phi}_n(x_0)$.

For computational tractability, we need the equivalent Conditional Flow Matching (CFM) objective 
$\mathcal{L}_{\text{CFM}}(\theta) = \mathbb{E}_{t,q(x_1),p_t(x|x_1)} \left\| v_t(x) - u_t(x|x_1) \right\|^2$ \citep{lipman_flow_2022}. Reparameterizing $p_t(x|x_1)$ in terms of just $x_0$ we get
\( \mathcal{L}_{\text{CFM}} = \mathbb{E}_{t, q(x_0, x_1)} \big{\|}  v_t  \left(\psi_t    (x_0 |x_1); \theta \right)  
    \ \ - u_t\left(\psi_t(x_0|x_1) | x_1 \right) \big{\|}^2, \)
where $\psi_t(x_0|x_1)$ is the conditional flow with a predefined form. We typically use
$ \psi_t(x_0|x_1) = (1 - t)x_0 + tx_1 $
with the corresponding vector field
$ u_t\left(\psi_t(x_0|x_1) | x_1\right) = x_1 - x_0 $.
%
%
%
Thus we obtain the loss 
\begin{equation}\label{key: 8.13-1}
\begin{aligned}
	\mathcal{L}_{\text{CFM}} = \mathbb{E}_{t, q(x_0, x_1)} \big{\|} x_0 + v_{t}(t, & \psi_t(x_0|x_1); \theta) \\
     & \qquad \quad -   \ x_1  \big{\|}^2.
\end{aligned}
\end{equation}

\subsection{The Training-Inference Gap }
Loss (\ref{key: 8.13-1}) can be seen as measuring the ground truth $x_1$ and the numerical result by implicit one-step Euler method within time interval $[0,1]$, with derivative estimated in time $t$.
We can also turn to a more advanced solving scheme, as we do during the inference phase. 
The difference is that when in training we use the ground truth value $\psi_t(x_0|x_1)$ since we have predefined the path, but when inferring we should use the estimated value $\hat{\phi}_n(x_0)$.  
And this creates the gap between the training stage and inference (or prediction). 
Fortunately, we can bound this gap by the following theorem (proof in Appendix \ref{appSec:proof}.). 
%
%

\begin{theorem}[Training-Inference Gap]\label{th:errBound}
	Assume that the truth vector field $u_t(x)$ is a Lipschitz-continuous
	function with the Lipschitz constant $L_u > 0$. 
	And the discrepancy between the learned vector field and the truth satisfies 
    $\mathbb{E}_{p(x_0)} \|v_{\theta}(t,\psi_t(x_0))-u_t(\hat{\phi}_t(x_0)) \|_{\infty} \le \delta $,
    then we can derive the following error estimate between the ground-truth values and the network's inferred values 
	\begin{equation*}
		\mathbb{E}\, |\varepsilon_{N}| \le \exp{(L_u t_{N-1})} \left( \Sigma_{j=0}^{N-1} (\delta \tau_j + \frac{1}{2}M\tau_j^2) + \mathbb{E}\, |\varepsilon_{0}| \right)	,
	\end{equation*} 
	where $M = \max\limits_{0 \leq t \leq 1} |\ddot{\psi_t}|$ is an upper bound for the second time derivative.
	Under special circumstances, when uniform step sizes are adopted, we obtain a more refined estimation,
	\begin{equation*}
		\begin{aligned}
			\mathbb{E}\, |\varepsilon_N| \leq \exp({L_u })  \, \mathbb{E}\, |\varepsilon_0|  + \frac{M\tau_0+2\delta}{2L}(\exp{(L_u })-1).
		\end{aligned}
	\end{equation*}
\end{theorem}

\begin{remark}[Theoretical Significance and Connections]
    This theorem serves as a diagnostic tool, offering an end-to-end decomposition of the inference error into components attributable to the training loss, the vector field network's properties ($L_u$), and numerical discretization. By explicitly quantifying the training-inference gap along the entire pipeline, it directly motivates our subsequent work on optimizing based on inference error (\ref{eq:mle-loss}). 
    Compared to \citet{benton2024error}, our work further accounts for the accuracy of numerical methods---a factor that significantly affects the model's practical inference performance, as demonstrated in Figures~\ref{fig:Results_different_numerical_precisions}, \ref{fig:numeric_error_comparison_heatmaps}, and \ref{fig:numeric_error_comparison_heatmaps_detailed}.
    Approximating the underlying vector field (e.g., through learning) introduces inherent errors. The resulting question of stability and error amplification for classical numerical methods is the core theoretical problem that our theorem resolves. 
    While \citet{zhouerror} also conduct error analysis, it is more focused on statistical learning theory, revealing the relationship between convergence rate and sample size. It does provide some numerical error analysis, but this belongs to the category of pure numerical analysis, quantifying the isolated discretization error of the approximate network itself rather than offering an end-to-end analysis from training loss to inference error. Its primary purpose is to guide how to balance computational cost and accuracy during inference.
\end{remark}
Though this shows that the gap between training and inference is bounded, such a gap will inevitably compromise the model's effectiveness.
This theorem indicates that the error in the final generated sample (or action in robot policy) $\varepsilon_N$ will further amplify the training error $\delta$, at least by a multiplicative factor $\exp({L_u })$ with an additive constant.
This issue becomes particularly severe when predefined paths may cause discontinuities in the vector field (Figure \ref{fig:flow_dis}), causing $L_u \to \infty$.
This motivates our pursuit of a consistent training-inference paradigm that can further optimaize $\varepsilon_N$. 
More specifically, we will utilize MLE for fine-tuning based on reconstruction results in the next section. 
\begin{remark}[Discontinuity]
    Discontinuity differs between FM (ODE) and diffusion models (SDE): SDE discontinuity stems from its (Brownian) diffusion term following statistical laws, which does not exhibit stiff behavior.
\end{remark}

\section{MLE Fine-Tuning Framework for FM}\label{sec:method}
In this section, we will introduce the specific methods for fine-tuning.
We first introduce the basic fine-tuning framework, and then we investigate a way to incorporate more designs into this framework to enhance its robustness.
\subsection{The Basic Framework}
We begin by making the following assumption regarding the conditional distribution of given samples.
\begin{assumption}\label{asmp:mle}
	Suppose that given the sample $x_1$, the underlying conditional distribution is a Gaussian distribution $p_{1}(x|x_1)=\mathcal{N}(x|x_1, \Sigma)$, where $\Sigma$ is a d-dimensional covariance matrix.
\end{assumption}
This Gaussian posterior assumption is grounded in the maximum entropy principle \cite{cover1999elements}, and its asymptotic validity is supported by the consistency theorem \cite{wied2012consistency}.
The following theorem specifies the MLE fine-tuning loss (proof and generalization in Appendix \ref{appSec:proof}). 
\begin{theorem}[MLE Training]\label{th:mle}
	Under Assumption \ref{asmp:mle}, and when $\Sigma$ is a scalar matrix, performing maximum likelihood estimation (MLE) by maximizing the expectation
	$\mathbb{E}_{q(x_0, x_1)} \left[\log  p_{1}\left(\hat{\phi}_N(x_0)|x_1\right) \right]$ is equivalent to minimizing the following loss function
	\begin{equation}\label{eq:mle-loss}
		\mathcal{L}_{\mathrm{MLE}} = \mathbb{E}_{q(x_0, x_1)} \left[ \| \varepsilon_N (x_0|x_1) \|^2 \right],
	\end{equation}
	where  $\varepsilon_n(x_0|x_1):=\psi_{t_n}(x_0|x_1)-\hat{\phi}_n(x_0)$ is the conditional model error. 
\end{theorem}
This forms the basic framework for fine-tuning. 
We only need to compute the model output $\hat{\phi}_N(x_0)$, and then optimize the relevant parameters using (\ref{eq:mle-loss}).
The detailed procedure is in Algorithm \ref{alg:flow_finetune}.
This enables the direct optimization of the model error, standing in contrast to previous approaches (Theorem \ref{th:errBound}), which could only provide a loose upper bound that contains non-optimizable components.

MLE by (\ref{eq:mle-loss}) has the advantange of high-precision for directly optimizing inference results. 
But it suffers from several critical issues. 
First, it is particularly prone to overfitting. Since no additional constraints are imposed on the vector field or trajectory shapes, it often generates sophisticated flow fields with convoluted solution paths \citep{finlay2020train}, reducing the model's reliability. 
Secondly, it is computationally expensive compared to the original FM training algorithm, for it needs repeated simulation of the ODE.

\textbf{However, we contend that MLE is particularly well-suited for fine-tuning pre-trained flow models}. The reasons are listed as follows. After the training of FM, we already get a relatively straight base model, MLE method will only fine-tune it. Thus, the vector fields will improves accuracy without significant shape distortion. From an optimization perspective, it more readily converges to local optima near a 'straight flow'. Moreover, this convergence process of parameters is significantly faster and more efficient than training a flow from scratch. Thus the higher computational complexity of MLE is acceptable. We emphasize that the complexity during training is generally inconsequential, as our primary focus remains inference speed which is unaffected by these training-phase design choices.

%

Also we want to clarify fine-tuning a flow model will not compromise its 'straighter' trajectory property and thus increase computational complexity. 
First, the trajectories FM are not always straight, since straighter lines between sample points does not necessarily mean that the path between distributions will also be straighter \citep{gao2024diffusion}. 
Second, the stochastic gradient descent optimization helps maintain the regularity of the vector field to some extent, avoiding the pathological stiffness that can arise from excessive straightness.
This will also be demonstrated in subsequent experiments. In this sense, FM optimizes a module (the vector field) by constructing a straight-line path, but practical constraints cause deviations from linearity. Therefore, a final optimization step can be applied directly to the reconstruction results for further performance enhancement.

We find in subsequent experiments that MLE fine-tuning is particularly effective in settings where the generated variables are of relatively low dimensionality yet exhibit highly complex dependencies on environmental conditions, such as those involving multimodal inputs.
Typical examples include policy generation and probabilistic numerical tasks
In contrast, image generation is considerably more challenging due to its high dimensionality and strong inter-pixel dependencies. Simulating such a complex ODE system is difficult, which poses substantial challenges for training.

\begin{figure*}[ht]
    \centering
    \begin{subfigure}[b]{0.23\linewidth}
        \centering
        \raisebox{1ex}{\includegraphics[width=\linewidth]{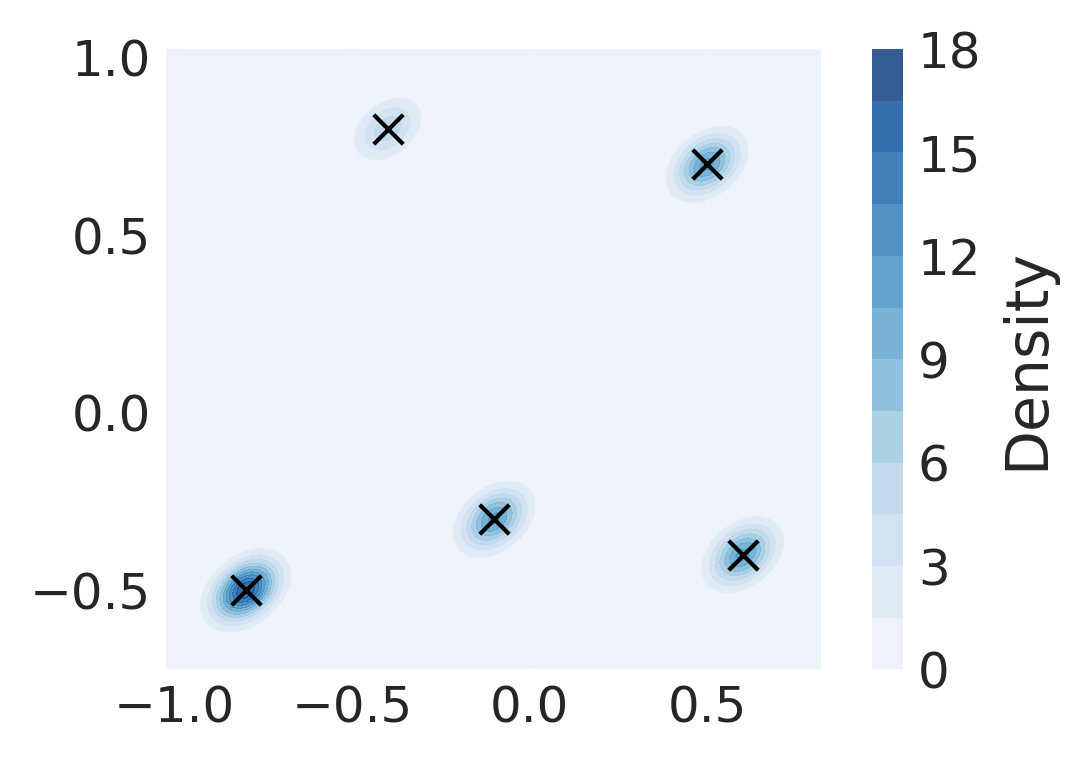}}
        \caption{Target Distribution.}
        \label{fig:target_distribution}
    \end{subfigure}
    \hfill
    \begin{subfigure}[b]{0.25\linewidth}
        \centering
        \includegraphics[width=\linewidth]{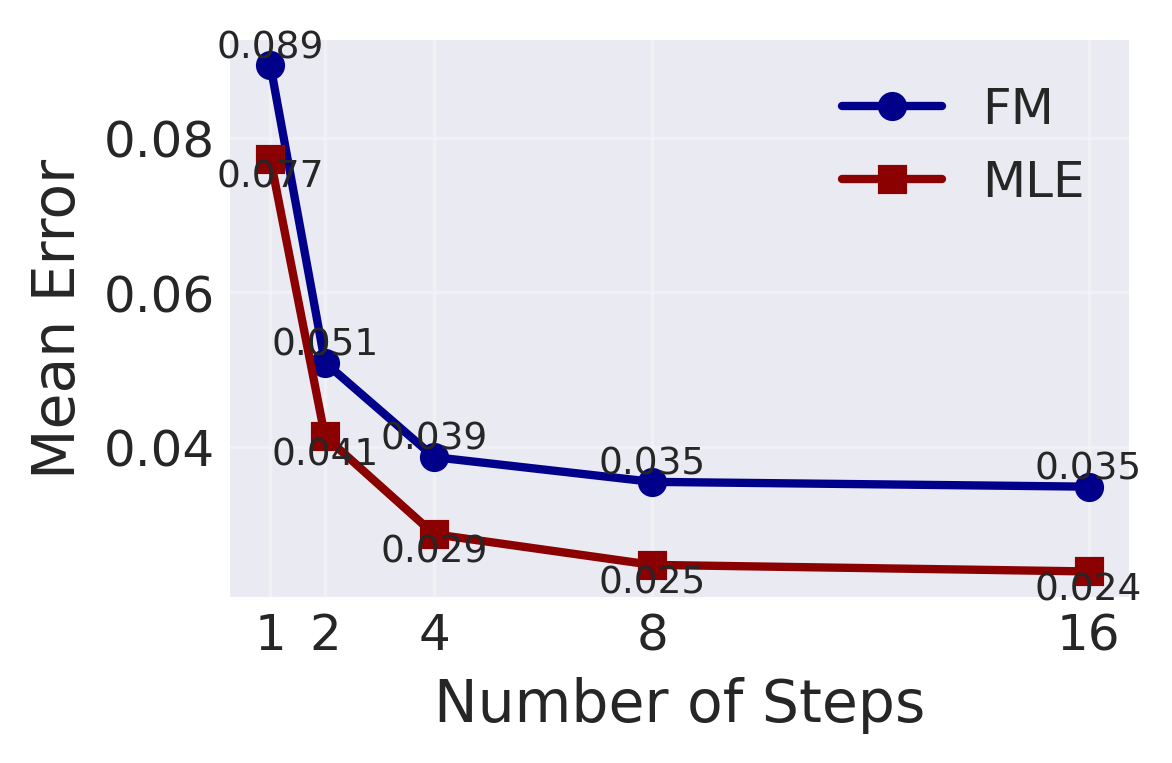}
        \caption{Error curves.}
        \label{fig:error_curves}
    \end{subfigure}
    \hfill
    \begin{subfigure}[b]{0.25\linewidth}
        \centering
        \includegraphics[width=\linewidth]{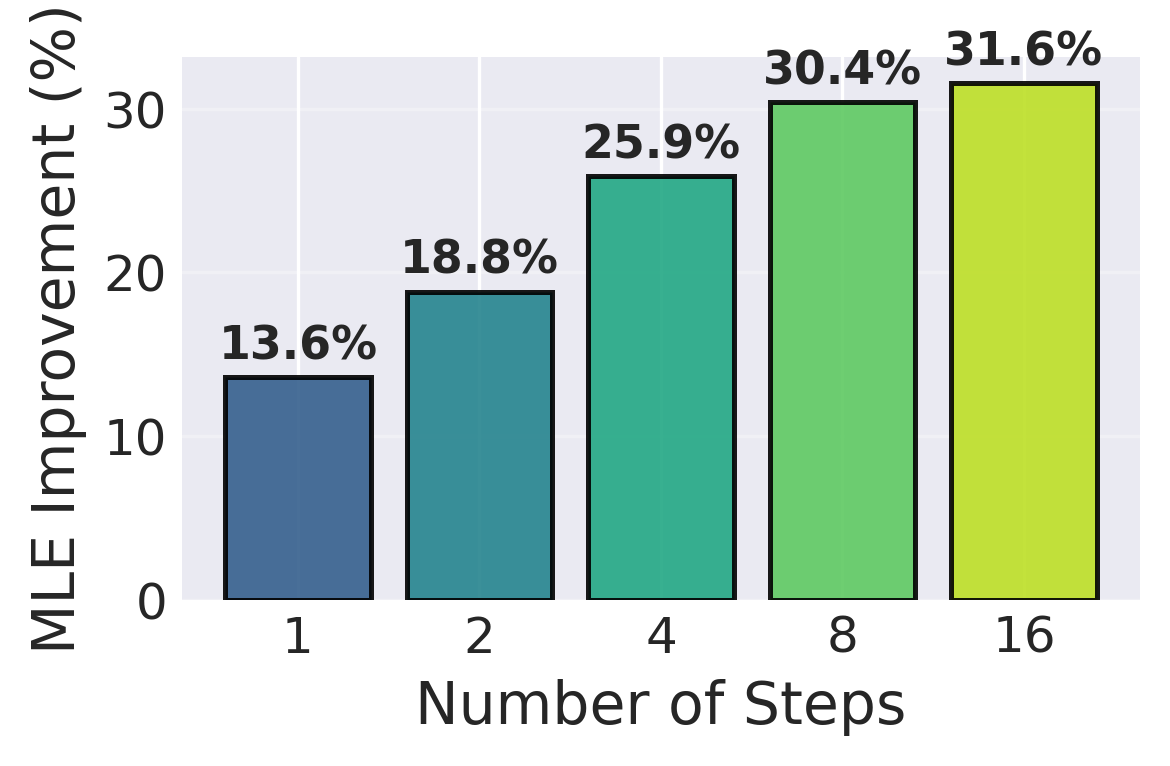}
        \caption{Relative improvement rate.}
        \label{fig:improvement_rate}
    \end{subfigure}
    \hfill
    \begin{subfigure}[b]{0.19\linewidth}
        \centering
        \raisebox{1ex}{\includegraphics[ width=\linewidth]{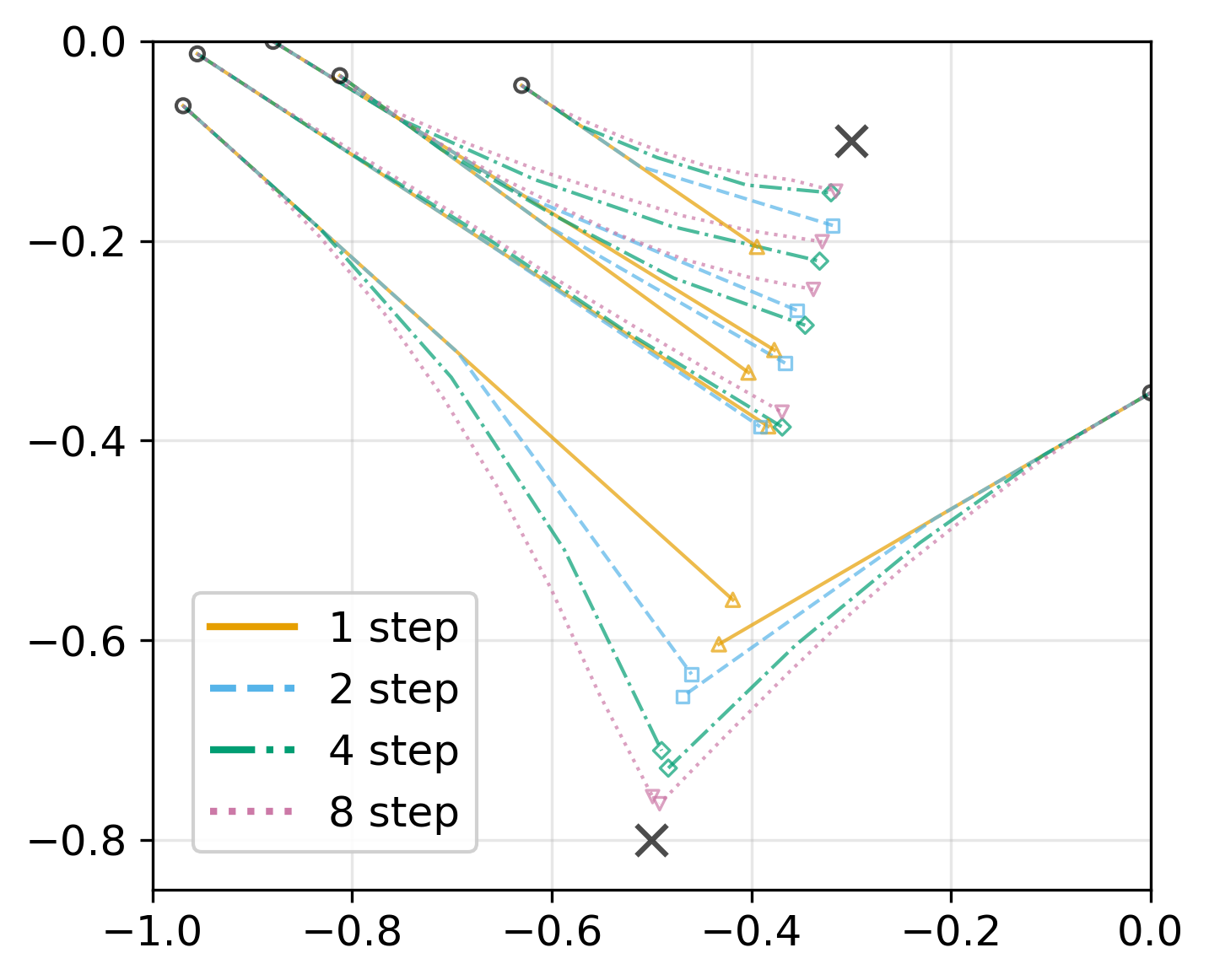}}
        \caption{Trajectory.}
        \label{fig:trajectories}
    \end{subfigure}
    \caption{Target Distribution and results under different numerical precisions.}
    \label{fig:Results_different_numerical_precisions}
\end{figure*}

\begin{figure*}[h]
	\centering
	\includegraphics[width=0.9\textwidth]{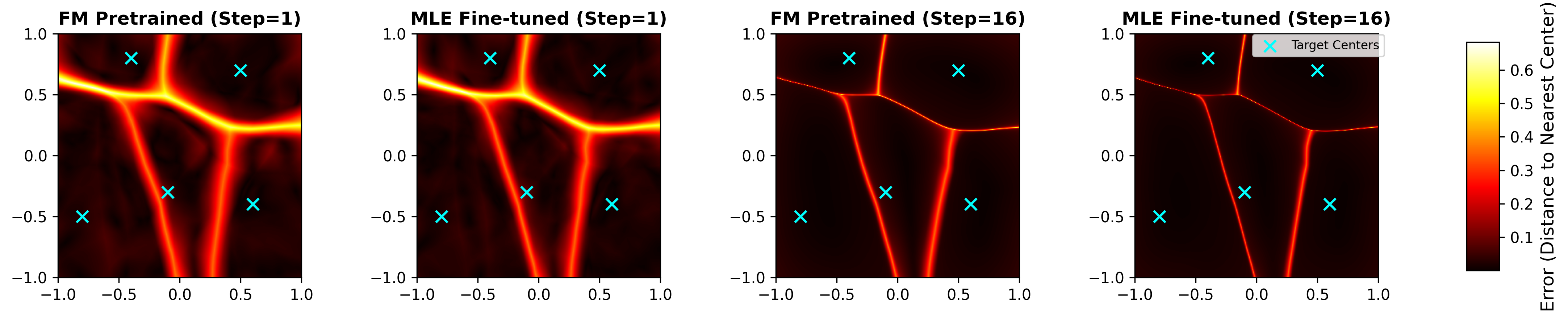}
	\caption{Error visualization across the latent space: a comparison of pre-trained and fine-tuned models at different numerical precisions. 
    }
	\label{fig:numeric_error_comparison_heatmaps}
\end{figure*}


\begin{figure*}[htbp]
    \centering
    \begin{subfigure}[b]{0.25\linewidth}
        \centering
        \includegraphics[width=\linewidth]{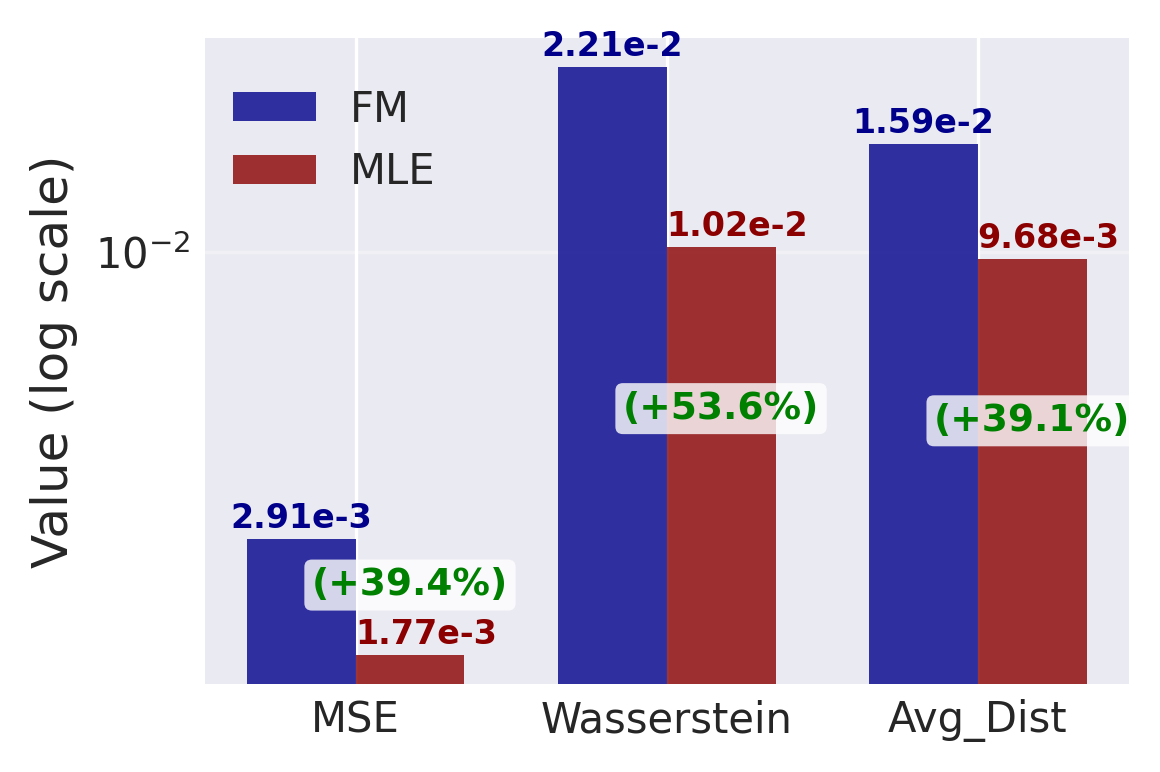}
        \caption{Quantitative Metrics} 
        \label{fig:quantitative_metrics} 
    \end{subfigure}
    \hfill 
    \begin{subfigure}[b]{0.25\linewidth}
        \centering
        \includegraphics[width=\linewidth]{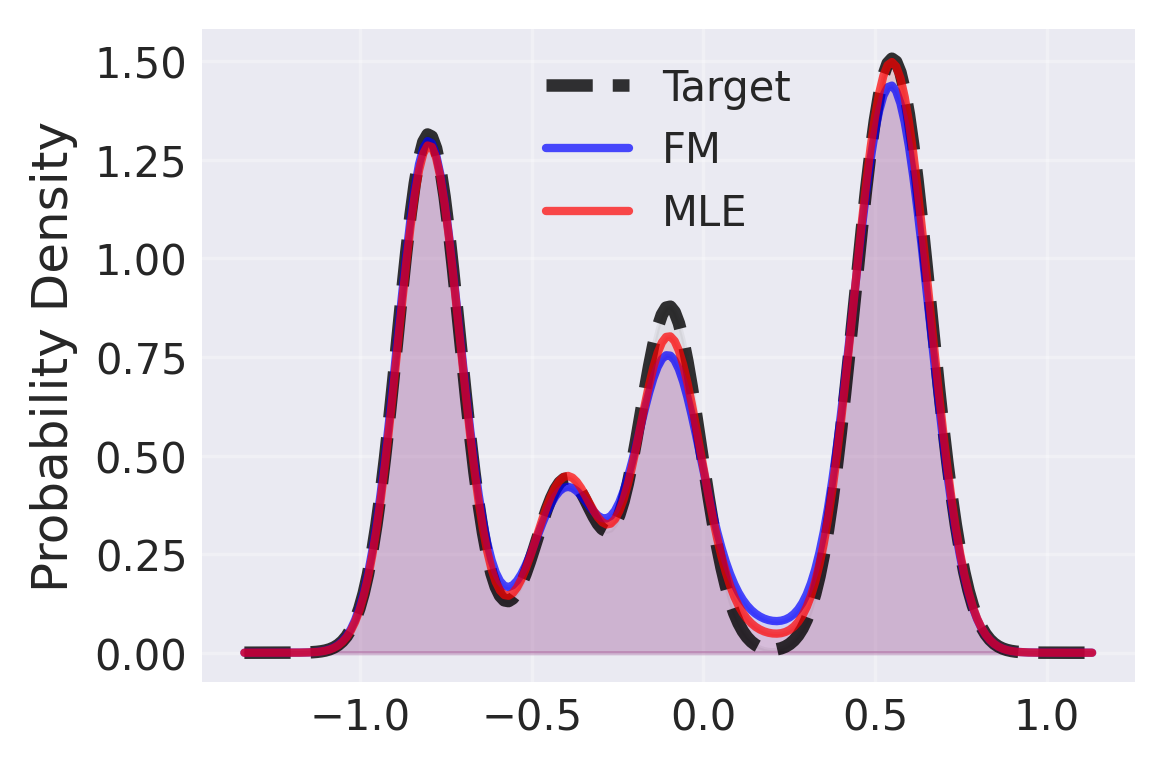}
        \caption{X-axis Distribution Density}
        \label{fig:x_axis_kde}
    \end{subfigure}
    \hfill
    \begin{subfigure}[b]{0.196\linewidth}
        \centering
        \includegraphics[width=\linewidth]{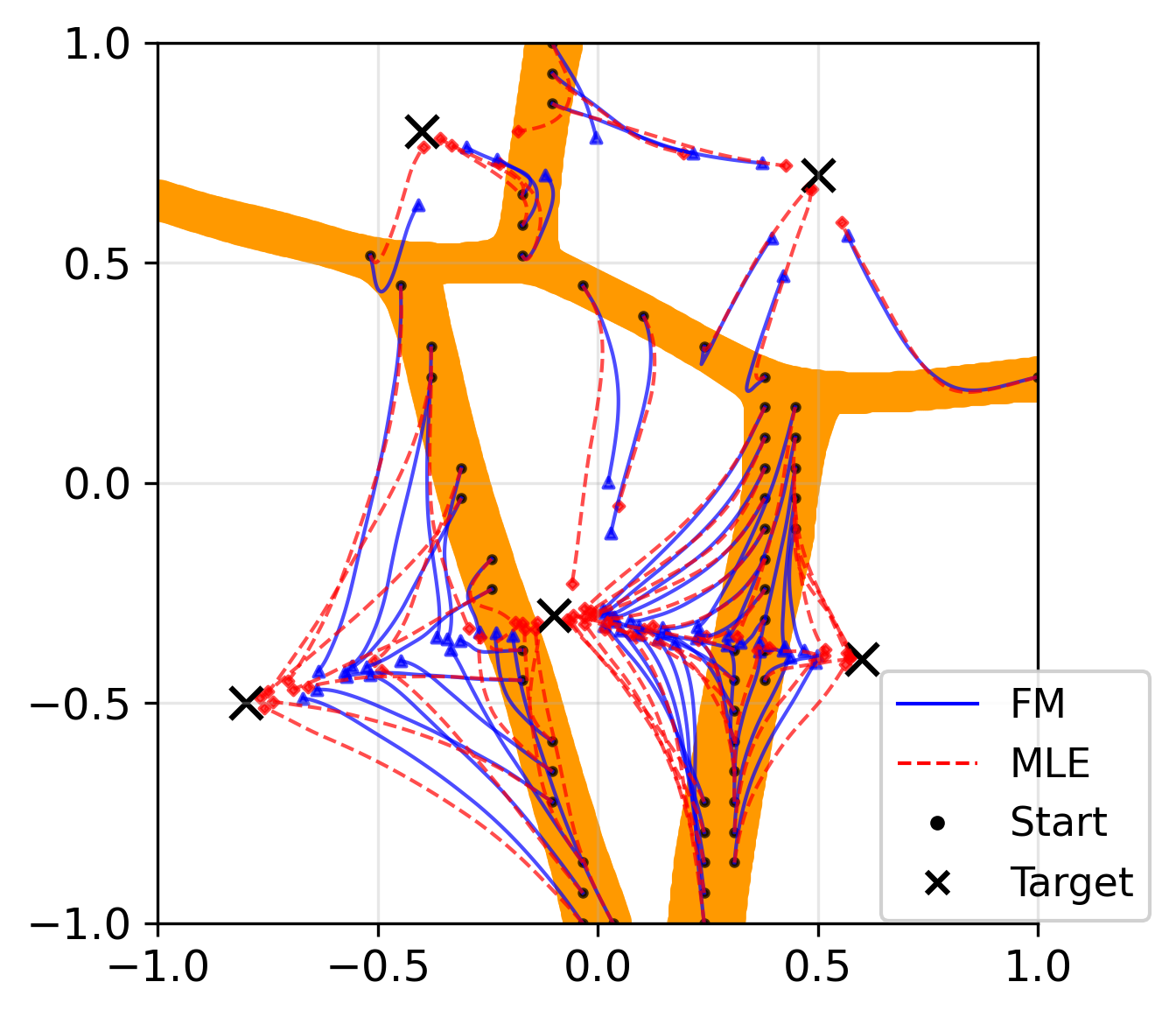}
        \caption{Trajectory Comparison}
        \label{fig:trajectory_comparison}
    \end{subfigure}
    \hfill
    \begin{subfigure}[b]{0.25\linewidth}
        \centering
       \raisebox{0.8ex}{\includegraphics[width=\linewidth]{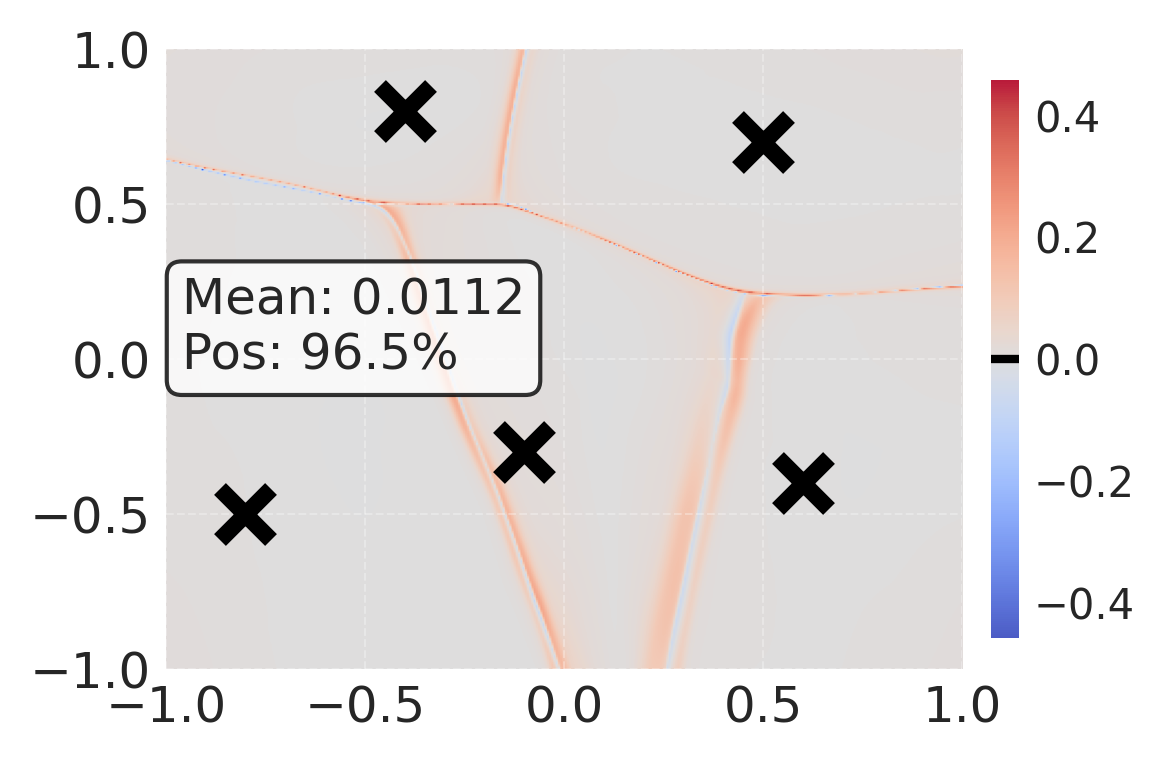}}
        \caption{Absolute Improvement}
        \label{fig:trajectory_improvement}
    \end{subfigure}
    \caption{Comparative Results of Pre-trained vs. Fine-tuned Models.}
    \label{fig:toy_results}
\end{figure*}

\subsection{Inducing Favorable Flow Properties}\label{sec:ResFT}


The MLE training paradigm relaxes direct constraints on the vector field, thereby enhancing its representational capacity. In this subsection, we futher illustrate that the MLE fine-tuning framework provides the flexibility to integrate further designs, thereby training flows with enhanced properties.


Specifically, we desire a flow that exhibits some stable and robust properties, such as 
Input-to-State Stability (ISS) and, furthermore, contraction (Figures \ref{fig:flow_c}-\ref{fig:flow_con_2dim}).
The precise definitions are provided in Appendix \ref{app:N_D_iss-con}.

\paragraph{Method: Generalized Artificial Viscosity.}
Viscosity is a mathematical expression describing a system’s resistance to deformation, acting as a damping force to dissipate energy and stabilize motion. 
When simulating physical systems which themselves may be non-dissipative, it is common practice in Scientific Computing to introduce artificial viscosity for stabilization \cite{vonneumann1950method,wilkins1980use,caramana1998formulations,margolin2023artificial}.
\citet{coutinho2023physics} recently experimented with learnable viscosity.
Viscosity can be viewed as analogous to negative feedback with a specific structure (details in Appendix \ref{app:viscous}).
Inspired by this, we introduce a generalized artificial viscosity (GAV) term, denoted by $\delta(\cdot)$, into FM to achieve more stable generation performance.

Here we choose $\delta (x)=A_0 x + \sum_{j=1}^M A_j f_j(W_j x)$ \cite{mei2024controlsynth}. To avoid cancellation effects from jointly optimizing the viscous and nominal terms, we employ a residual learning framework \cite{yuanpolicy,jiang2025transic}.
Denote $\tilde{v}$ the vector field of the residual part with parameters $\tilde{\theta}$. Therefore,
\begin{equation}\label{eq:sys_delta}
    \frac{d}{dt} \phi_t  = \tilde{v}_t(\bm o; \tilde{\theta})+\delta (\phi_t),\  1\le t \le 2,
\end{equation}
with initial $\phi_1(x_0)$ and output $\phi_2(x_0)$. We then have:
\begin{theorem}[ISS and Contraction]\label{thm:iss-con}
    Suppose the nonlinearities $f_j^i$ are all continuous, monotonically increasing, and passing through the origin (see Assumptions \ref{asmp:sign}-\ref{asmp:con-inc}). When the weights $A_j,W_j$ in $\delta(\cdot)$ satisfy certain linear matrix inequalities (LMIs) (see (\ref{eq:ISS-condition}) and (\ref{eq:con-condition})), then system (\ref{eq:sys_delta}) is ISS or contracting.
\end{theorem}
Details and proofs in Appendix \ref{app:iss-con}.
Theorem \ref{thm:iss-con} requires satisfing LMIs simultaneously with optimizing parameters via loss (\ref{eq:mle-loss}), which poses a practical challenge. 
So we further investigate a direct parameterization, inspired by \citet{wang2023direct,revay2023recurrent}, to satisfy these LMIs naturally during optimization.
\begin{corollary}[Direct Parameterization]\label{cor:thm}
Let $\delta(x)=A_0 x(t) + \sum_{j=1}^M A_j f_j(D_j x(t))$ where $D_j$ are all positive diagonal matrices and $A_j$ are all Hurwitz, i.e., all eigenvalues have strictly negative real parts. Then system (\ref{eq:sys_delta}) is contracting.
\end{corollary}
Proof in Appendix \ref{app:iss-con-cor}. We can use Schur decomposition and Cayley transform to construct a Hurwitz matrix mentioned (details in Appendix \ref{app:iss-con-cor}). Thus, we can get a direct parameterization scheme (specific steps in Algorithm \ref{alg:GAV}).

\section{Experiments}\label{experiments}
In this section we experimentally evaluate the benefits of fine-tuning FM.
First we design a toy example to verify the basic principles of MLE fine-tuning. 
We then conduct experiments on complex real-world tasks, including policy generation and spatial extrapolation.
Image experiments that involve higher-dimensional generative variables and all experiment details are in Appendix~\ref{appSec:exp-de}. The source code is provided in the supplementary material.

\subsection{Toy Example}\label{sec:toy_example}

Suppose there are several target centers; we aim to learn the underlying distribution and infer the number and locations of these centers from scattered sample points contaminated by additive white Gaussian noise. 
This formulation captures a broad and important class of real-world tasks, with concrete examples detailed in Appendix \ref{app:exp_toy}.
We employ a Gaussian mixture model as the target distribution, as shown in Figure~\ref{fig:target_distribution}.

\paragraph{Training Configuration.}
We trained a FM model for 2,000 epochs (A100 GPU: $57.3s$) to ensure convergence. 
We then performed MLE fine-tuning for 1000 epochs from the 1000th-epoch checkpoint (A100 GPU: $28.7s$ pre-training, $23.2s$ fine-tuning, $51.9s$ total).
The hyperparameter values, such as the learning rate and batch size, are exactly the same.
The Minibatch Optimal Transport \cite{tong2024improving} technique is applied.

\paragraph{Results and Analysis.}
The performance of the pre-trained FM and MLE fine-tuned models was evaluated and results are shown in Figures~\ref{fig:Results_different_numerical_precisions}-\ref{fig:toy_results}. 
Figure~\ref{fig:Results_different_numerical_precisions} plots the performance of models under different numerical discretization steps. 
The metric is chosen as the Euclidean distance to the nearest Gaussian center.
Figure \ref{fig:trajectories} discretizes the space within the range $[-1, 0]^2$ into a $10\times10$ grid as different initial points and plots the trajectories whose terminal error exceeds $0.1$. 
Figure~\ref{fig:numeric_error_comparison_heatmaps} visually presents the inference error for different initial points in the latent space. 
Figure~\ref{fig:toy_results} provides a comprehensive comparison between the pre-trained and fine-tuned models across multiple aspects. These include various performance metrics (a), probability density functions (b), a visual comparison of trajectories (c), and the improvement magnitude across different regions of the latent space (d).Figure \ref{fig:trajectory_comparison} discretizes the space $[-1, 1]^2$ into a 30×30 grid and plots the trajectories originating from all initial points where the inference error exceeds 0.1. When the space is discretized more finely into a $360\times 360$ grid, all regions containing such initial points (with error $> 0.1$) are highlighted in yellow.
Taken together, these visualizations support the following findings.

\textbf{\textit{Overall Comparison.}} MLE fine-tuning enhances model performance (Figure~\ref{fig:toy_results}), including changes in different metrics, probability density functions, and trajectories with endpoints closer to the centers.

\textbf{\textit{Impact of Numerical Precision.}} Numerical precision significantly influences model performance (Figures~\ref{fig:Results_different_numerical_precisions}-\ref{fig:numeric_error_comparison_heatmaps}). And MLE fine-tuning consistently achieves better results across all precision scales.

\textbf{\textit{Inference Trajectory.}} The straight paths constructed for training in FM do not result in purely linear trajectories during actual inference (Figure~\ref{fig:error_curves}). Moreover, MLE fine-tuning brings the endpoints closer to the sample centers without increasing the trajectory’s curvature (Figure~\ref{fig:trajectory_comparison}), thereby avoiding additional inference difficulty.

\textbf{\textit{Error distribution.}} 
The inference error exhibits a pronounced peak precisely at the boundaries that demarcate the attracting regions of distinct latent modes, in contrast to the relatively lower error observed within each basin (Figure~\ref{fig:numeric_error_comparison_heatmaps}).
This is consistent with our analysis, especially under the straight-path assumption, because the vector field exhibits its most pathological behavior—even potential discontinuity—precisely at the boundary points.
The latter two subfigures in Figure~\ref{fig:numeric_error_comparison_heatmaps} demonstrate that MLE fine-tuning alleviates the issue by relaxing constraints on the vector field. A key visual indicator is the thinning of the red boundary lines. This finding is further corroborated by  Figure~\ref{fig:trajectory_improvement}, which shows that the improvement is most pronounced within the boundary regions. 

These results not only empirically validate the superiority of MLE fine-tuning in performance, but also corroborate our earlier in-depth analysis regarding the inherent limitations of FM training and the necessity of MLE fine-tuning.

\begin{table*}[t]
	\centering
	\resizebox{\linewidth}{!}{
\begin{tabular}{l|cccc|cccc|cccc}
    \toprule[1.0pt]
    \multirow{2}{*}{\textbf{Methods }} & \multicolumn{4}{c|}{\textbf{Franka Kitchen }} & \multicolumn{4}{c|}{\textbf{Push-T} } & \multicolumn{4}{c}{\textbf{Robomimic }} \\
    \cmidrule(lr){2-5} \cmidrule(lr){6-9} \cmidrule(lr){10-13}
    & Converged & Early & Noise(G) & Noise(S)
    & Converged & Early & Noise(G) & Noise(S)
    & Converged & Early & Noise(G) & Noise(S)
      \\
    \midrule 
    DDPM (100-step) & 0.771 \tiny{ $\pm$ 0.01} & 0.395\tiny{ $\pm$ 0.03} & 0.611\tiny{ $\pm$ 0.03} & 0.523 \tiny{ $\pm$ 0.03}& 0.717\tiny{ $\pm$ 0.01}& 0.510\tiny{ $\pm$ 0.02} & 0.603\tiny{ $\pm$ 0.02} &   0.534\tiny{ $\pm$ 0.01} & 0.756\tiny{ $\pm$ 0.02} & 0.322\tiny{ $\pm$ 0.04} &  0.542 \tiny{ $\pm$ 0.02}&0.501\tiny{ $\pm$ 0.03} \\
    DDPM (DPM sampling) & 0.751 \tiny{ $\pm$ 0.02}& 0.344 \tiny{ $\pm$ 0.03}& 0.592 \tiny{ $\pm$ 0.03}& 0.519 \tiny{ $\pm$ 0.03}&  0.653 \tiny{ $\pm$ 0.02} & 0.505 \tiny{ $\pm$ 0.03} & 0.577 \tiny{ $\pm$ 0.02} & 0.467 \tiny{ $\pm$ 0.02} & 0.711\tiny{ $\pm$ 0.02} & 0.252 \tiny{ $\pm$ 0.02} & 0.533 \tiny{ $\pm$ 0.03} & 0.480 \tiny{ $\pm$ 0.03} \\
    DDIM & 0.747 \tiny{ $\pm$ 0.02}& 0.317\tiny{ $\pm$ 0.02}  &0.585\tiny{ $\pm$ 0.02}  & 0.512 \tiny{ $\pm$ 0.02}  & 0.637\tiny{ $\pm$ 0.02} & 0.496\tiny{ $\pm$ 0.02} & 0.571\tiny{ $\pm$ 0.02} & 0.459\tiny{ $\pm$ 0.02} & 0.707\tiny{ $\pm$ 0.03} &0.227\tiny{ $\pm$ 0.03} & 0.512\tiny{ $\pm$ 0.02} & 0.477\tiny{ $\pm$ 0.02} \\
    \hline
    \rowcolor{lightpink} 
    FM & 0.742\tiny{ $\pm$ 0.02}& 0.378\tiny{ $\pm$ 0.04}  & 0.597\tiny{ $\pm$ 0.02}  & 0.515\tiny{ $\pm$ 0.03}   & 0.684\tiny{ $\pm$ 0.02} & 0.499\tiny{ $\pm$ 0.03} & 0.591\tiny{ $\pm$ 0.02} & 0.461\tiny{ $\pm$ 0.02} & 0.729\tiny{ $\pm$ 0.03} & 0.314\tiny{ $\pm$ 0.05} & 0.537\tiny{ $\pm$ 0.02} & 0.491\tiny{ $\pm$ 0.03} \\
    FM (SDE sampling) & 0.723\tiny{ $\pm$ 0.02} & 0.471\tiny{ $\pm$ 0.04} & 0.602\tiny{ $\pm$ 0.02} & 0.513\tiny{ $\pm$ 0.02} & 0.665\tiny{ $\pm$ 0.02} & 0.526\tiny{ $\pm$ 0.03} & 0.588\tiny{ $\pm$ 0.02} & 0.442\tiny{ $\pm$ 0.02} & 0.715\tiny{ $\pm$ 0.03} & 0.387\tiny{ $\pm$ 0.05} & 0.536\tiny{ $\pm$ 0.03} & 0.485\tiny{ $\pm$ 0.03} \\
    \rowcolor{lightblue}
    FM (Transformer) & 0.661\tiny{ $\pm$ 0.02} & 0.325 \tiny{ $\pm$ 0.05} & 0.519\tiny{ $\pm$ 0.02} & 0.442 \tiny{ $\pm$ 0.02} & 0.573 \tiny{ $\pm$ 0.02}& 0.334\tiny{ $\pm$ 0.04} & 0.331\tiny{ $\pm$ 0.02} & 0.307\tiny{ $\pm$ 0.02} &0.598\tiny{ $\pm$ 0.02} & 0.251\tiny{ $\pm$ 0.05} & 0.396\tiny{ $\pm$ 0.02} & 0.255\tiny{ $\pm$ 0.02} \\
    \hline
    \rowcolor{lightpink}
    \textbf{FT-FM}  & \textbf{0.811}\tiny{ $\pm$ 0.02}& \textbf{0.821} \tiny{ $\pm$ 0.02}& 0.655\tiny{ $\pm$ 0.02} & 0.530 \tiny{ $\pm$ 0.02}  & \textbf{0.751}\tiny{ $\pm$ 0.02}& \textbf{0.757} \tiny{ $\pm$ 0.02}& 0.622 \tiny{ $\pm$ 0.02}& 0.541 \tiny{ $\pm$ 0.02}& \textbf{0.791}\tiny{ $\pm$ 0.02} & \textbf{0.755}\tiny{ $\pm$ 0.02} & 0.592\tiny{ $\pm$ 0.02}& 0.517 \tiny{ $\pm$ 0.02} \\
    \textbf{FT-FM} (SDE sampling) & 0.782 \tiny{ $\pm$ 0.02}& 0.791 \tiny{ $\pm$ 0.02}& 0.676\tiny{ $\pm$ 0.03} &  0.518\tiny{ $\pm$ 0.03} & 0.744\tiny{ $\pm$ 0.02}  & 0.750\tiny{ $\pm$ 0.02} & 0.625\tiny{ $\pm$ 0.03} & 0.538\tiny{ $\pm$ 0.02} & 0.786\tiny{ $\pm$ 0.02} & 0.747\tiny{ $\pm$ 0.03} &  0.577\tiny{ $\pm$ 0.02} & 0.511\tiny{ $\pm$ 0.02} \\
     \rowcolor{lightblue}
    \textbf{FT-FM} (Transformer) & 0.712\tiny{ $\pm$ 0.02} & 0.638\tiny{ $\pm$ 0.02} & 0.544\tiny{ $\pm$ 0.02} & 0.472\tiny{ $\pm$ 0.02} & 0.645\tiny{ $\pm$ 0.02} & 0.633\tiny{ $\pm$ 0.03} & 0.509\tiny{ $\pm$ 0.02} & 0.431\tiny{ $\pm$ 0.02} & 0.651\tiny{ $\pm$ 0.02} & 0.599\tiny{ $\pm$ 0.02} & 0.421\tiny{ $\pm$ 0.02} & 0.301\tiny{ $\pm$ 0.02} \\
    \textbf{FT-FM} (GAV)& 0.784\tiny{ $\pm$ 0.01} & - & \textbf{0.704}\tiny{ $\pm$ 0.01} & \textbf{0.581}\tiny{ $\pm$ 0.02}   & 0.749\tiny{ $\pm$ 0.01} & - & \textbf{0.672}\tiny{ $\pm$ 0.02} & \textbf{0.582}\tiny{ $\pm$ 0.01} & 0.788 \tiny{ $\pm$ 0.01} & - & \textbf{0.642} \tiny{ $\pm$ 0.01} & \textbf{0.571} \tiny{ $\pm$ 0.02} \\
    \hline
    FT-FM (GAV, ablation) & 0.786\tiny{ $\pm$ 0.02} & - & 0.641\tiny{ $\pm$ 0.03} & 0.507\tiny{ $\pm$ 0.03}  & 0.746\tiny{ $\pm$ 0.02} & - & 0.606\tiny{ $\pm$ 0.02} & 0.517\tiny{ $\pm$ 0.02} &  0.790\tiny{ $\pm$ 0.02} & - & 0.551 \tiny{ $\pm$ 0.02} & 0.510 \tiny{ $\pm$ 0.03} \\		
    \bottomrule[1.0pt]
\end{tabular}
	}
	\caption{
		Model success rate (with standard deviation) in three scenarios. Our method appears in bold in the first column. 
        The ``converged'' (fully trained) and \enquote{early} (trained for one-third of epochs) columns report model performance and fine-tuning results using the corresponding models, respectively.
        The columns \enquote{Noise(G)} and \enquote{Noise(S)} correspond to inputs perturbed by Gaussian noise ($\sigma=0.01$) and salt-and-pepper noise (0.5\% pixel ratio), respectively.
	}
	\label{tab:robot-SR}
\end{table*}

\begin{table*}[t]
\begin{center}
\begin{scriptsize}
\begin{tabular}{lccccccccc}
\toprule
Model & CNP & NP & ANP & TNP & NDP & FLOWNP & FT-FLOWNP & FT-FLOWNP(GAV) & FT(ablation) \\
\midrule
Tar\_ll    & 3.66 {\tiny $\pm$ 0.016}    & 3.54 {\tiny $\pm$ 0.017} & 6.62 {\tiny $\pm$ 0.195} & 13.36 {\tiny $\pm$0.017} & 6.45 {\tiny $\pm$ 0.067}  & 17.98 {\tiny $\pm$ 0.017} & \textbf{18.31} {\tiny $\pm$ 0.022} &  \textbf{18.27} {\tiny $\pm$ 0.013} &  18.25 {\tiny $\pm$ 0.031} \\
\bottomrule
\end{tabular}
\end{scriptsize}
\end{center}
\caption{Target log-likelihood mean $\pm$ 1 standard deviation for the ERA5 meteorological dataset.}
\label{tab:era5}
\end{table*}

\begin{figure}[t]
    \centering
    \begin{subfigure}[b]{0.4\textwidth}
        \centering
        \includegraphics[width=\textwidth]{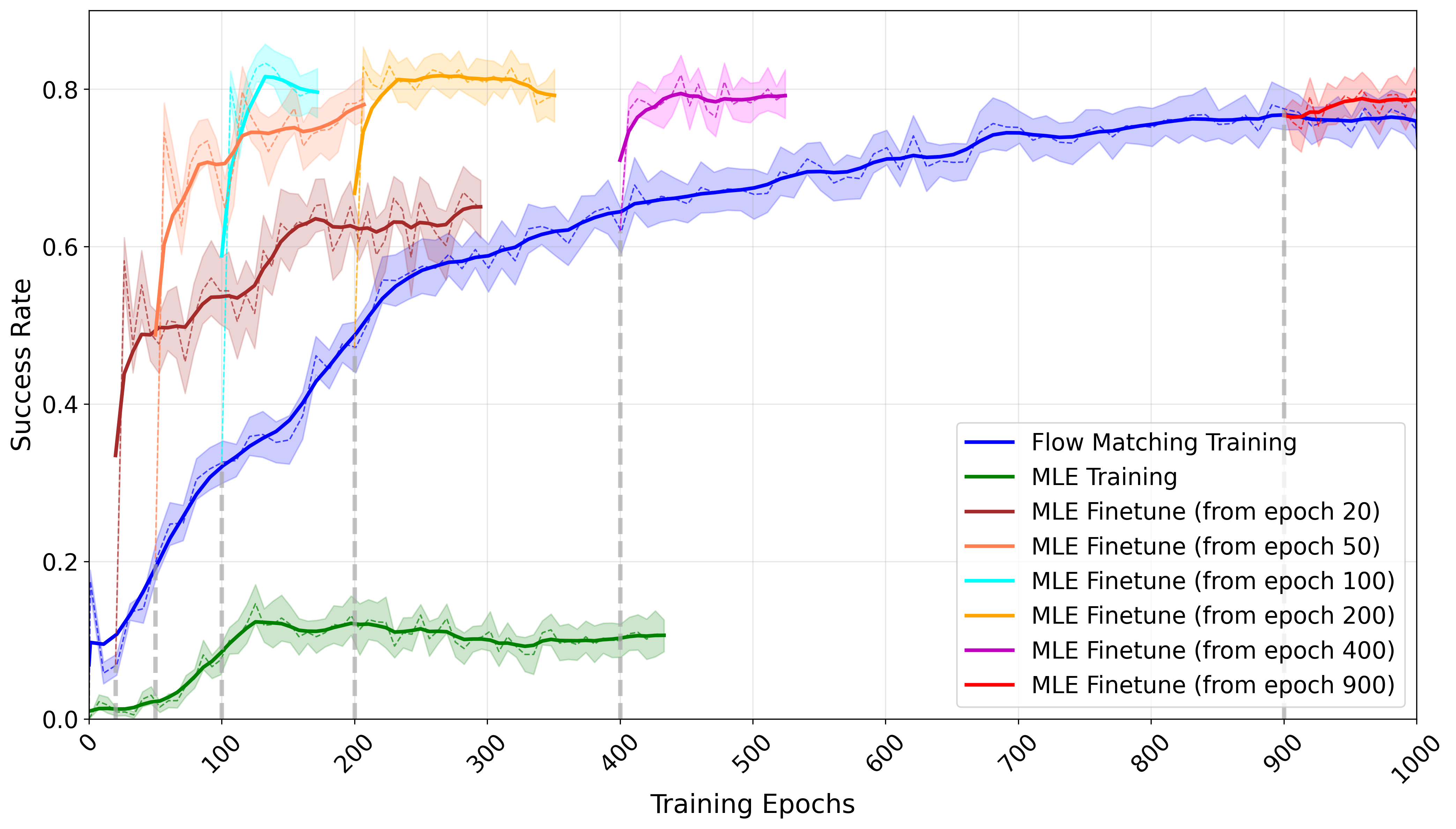}
        \caption{Success Rate in the Franka Kitchen Environment.}
        \label{fig:kitchen_sub}
    \end{subfigure}
    \hfill 
    \begin{subfigure}[b]{0.4\textwidth}
        \centering
        \includegraphics[width=\textwidth]{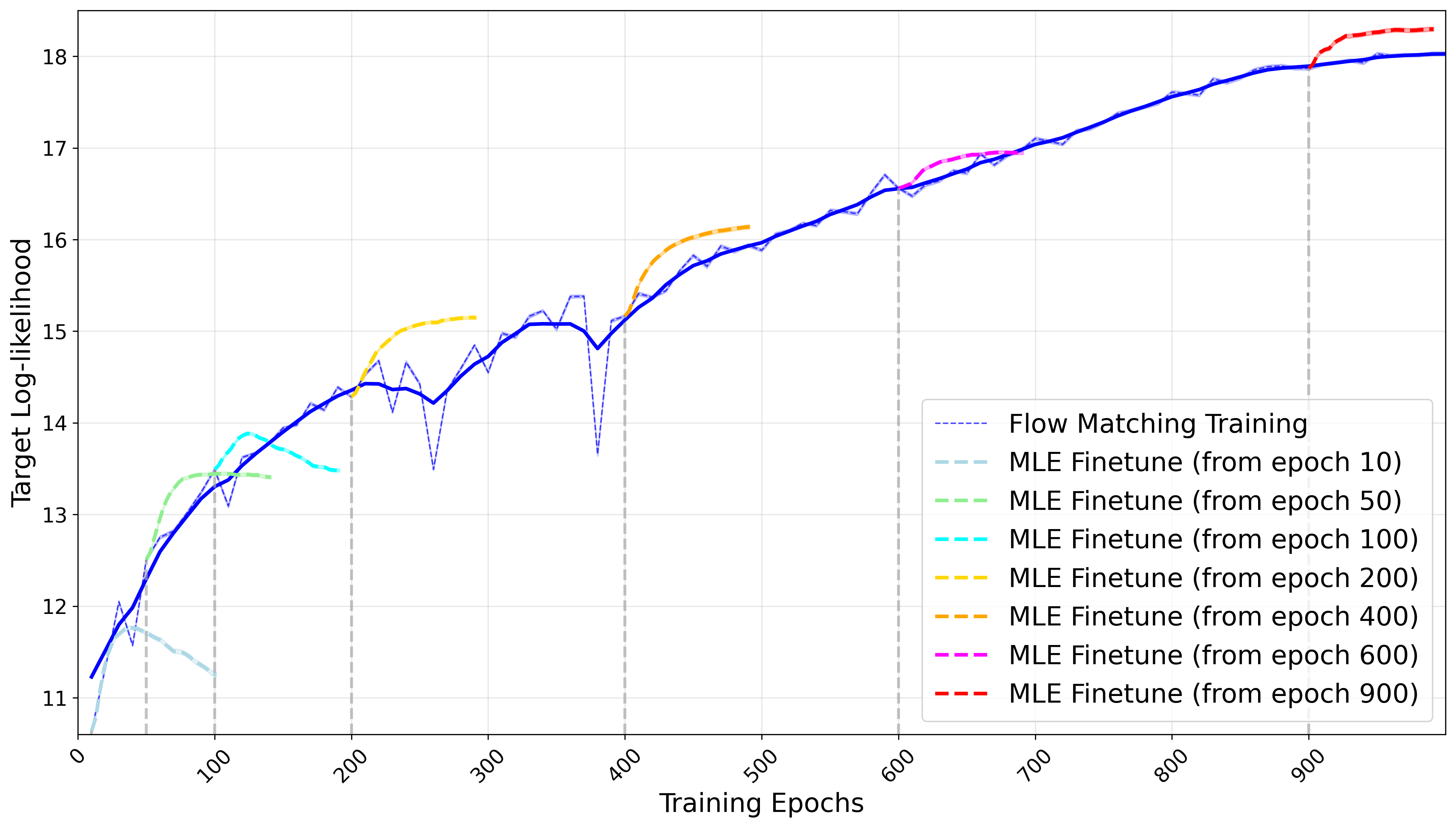}
        \caption{Target Log-likelihood on the ERA5 Dataset.}
        \label{fig:era5_sub}
    \end{subfigure}
    \caption{Evaluation of MLE fine-tuning at different stages. For the pre-trained FM curve, the horizontal axis represents the epoch count. For the fine-tuning curve, the horizontal axis is scaled accordingly to maintain the same time scale as pre-training. Light shading: standard deviation. Corresponding training error curves are in Figures \ref{fig:era5_curve_loss} and \ref{fig:kitchen_curve_loss}.}
    \label{fig:combined_results}
\end{figure}

\begin{figure}[t]
    \centering
    \includegraphics[width=0.45\textwidth]{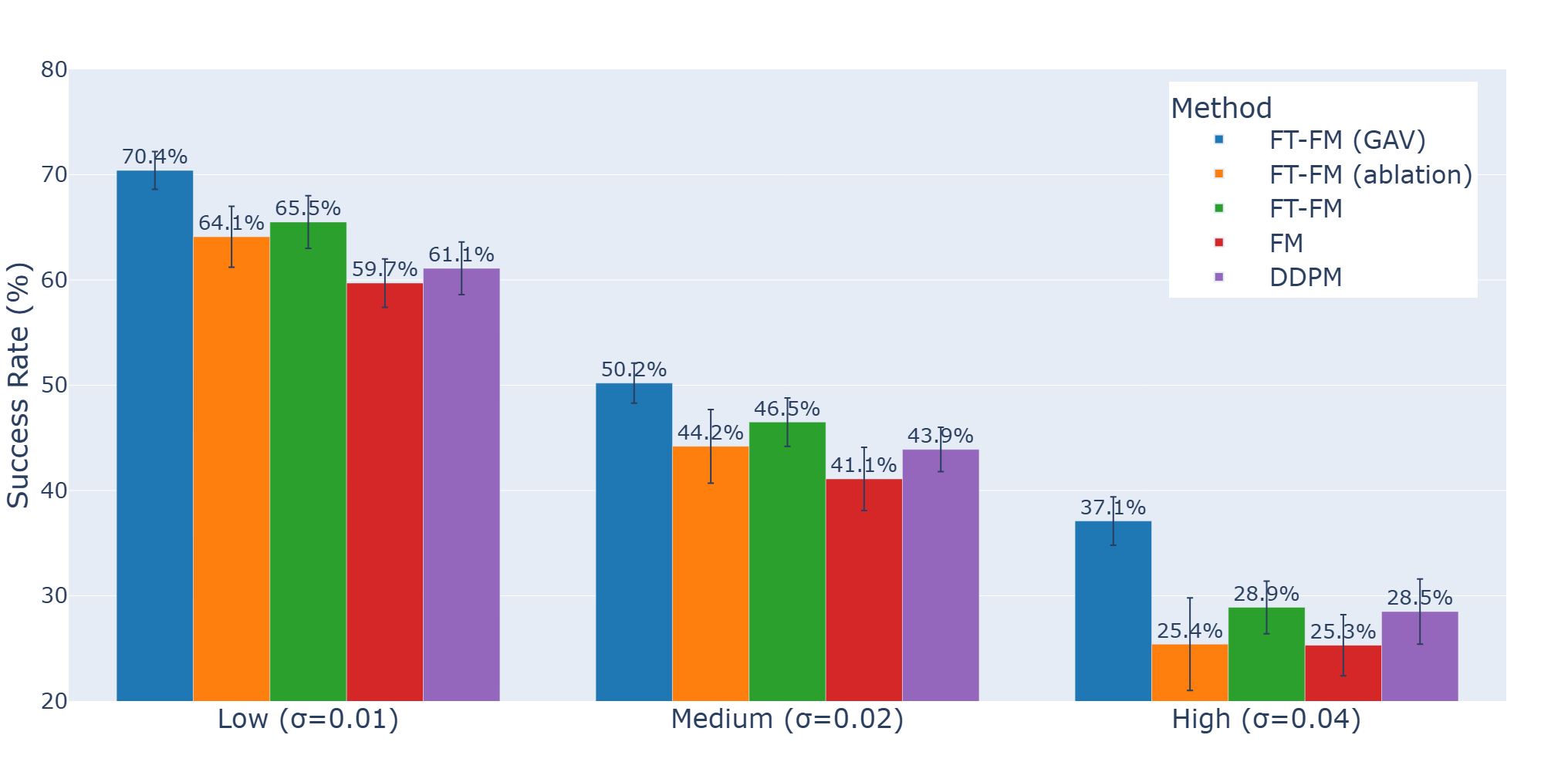}
    \caption{Success Rate under Different Gaussian Noise Levels (\%) in the Franka Kitchen Environment.}
    \label{fig:noise_GAV}
\end{figure}

\subsection{Robotic Manipulation} \label{sec:experiment_ro}

\paragraph{Description of Tasks and Datasets.}
We choose three benchmark robot manipulation datasets which includes closed-loop 6D robot actions and gripper actions: Franka Kitchen \citep{gupta2020relay}, push-T \citep{florence2022implicit}, and Robomimic \citep{mandlekar2022matters}. 
MuJoCo and Gym are used as the test environments
(visualization in Figure~\ref{fig:mujoco}).
\begin{itemize}[leftmargin=0.35cm, itemsep=0em, topsep=0em]
    \item
	The Franka Kitchen environment contains 7 interactive objects with 566 human demonstration sequences.
	Each demonstration completes any 4 tasks in variable order, and the goal is to accomplish as many tasks as possible regardless of sequence. 
	The policy uses state-based observations and generates closed-loop commands for both robot joint movements and gripper actions.
    
	\item 
	Push-T involves manipulating a T-shaped block to a designated target using a circular end-effector. The policy takes RGB images along with end-effector proprioception as input and produces closed-loop end-effector actions. Dataset includes 200 human demonstrations.
	
	\item
	Robomimic offers 5 different tasks with high-quality human teleoperation demonstrations. This study utilizes the Transport task, containing 200 demonstrations. The policy operates on state-based inputs and outputs closed-loop control signals for robot joints and the gripper.
\end{itemize}

\paragraph{Experimental Setup.}
We choose DDPM, DPM sampling \cite{lu2022dpm}, DDIM, FM, SDE sampling \cite{ma2024sit} as baseline models. Backbone defaults to U-Net, while Transformer version also tested. The evaluation in each environment has been carried out across 400 different initial conditions. The baseline models are trained for 4500 epochs.  
The fine-tuned training starts from the checkpoint at 3500 epochs and runs for approximately 100 epochs to ensure a roughly consistent training duration.
For ablation study, we replace the viscosity with a simple MLP.

\paragraph{Results and Analysis.}
The results are recorded in \autoref{tab:robot-SR} and Figures ~\ref{fig:kitchen_sub} and \ref{fig:noise_GAV}. 
From Figure~\ref{fig:kitchen_sub} we can see that MLE fine-tuning quickly improves model performance, regardless of the presence of noise or the choice of backbone (UNet or Transformer).
We also observe that fine-tuning does not require the pre-trained model to be fully converged; even early-stage MLE fine-tuning leads to significant performance gains.
A concrete example: after merely a few epochs of fine-tuning on a model at epoch 100, its success rate jumped dramatically from just above 30\% to over 80\%.
This benefit stems from the powerful fitting capacity of the MLE training paradigm, yet it also introduces some risk of overfitting. For instance, some fine-tuning curves exhibit a declining trend after reaching their performance peak.
Also MLE training from scratch is difficult, which consequently hinders significant performance improvement.
Table \ref{tab:robot-SR} and Figure \ref{fig:noise_GAV} demonstrate that introducing GAV can effectively enhance the model's robustness to noise.
Under Gaussian and salt-and-pepper noise, the GAV prevents rapid performance degradation of the model.
In contrast, replacing GAV with a simple MLP in the ablation study brings no robustness improvement.

\subsection{Meteorological Data}
\paragraph{Tasks, Datasets, and Experimental Setup.}
To assess the performance of MLE fintuning in real-world high-dimensional tasks, we use meteorological data from the ERA5 global dataset \cite{hersbach2020era5}.
We adopt the Flow Matching Neural Process (FlowNP) framework \cite{hamad2025flow} and follow the steup in \citet{hamad2025flow,holderrieth2021equivariant} and  to perform spatial extrapolation conditioned on random context points.
The data covers a circular region of 520km radius centered on Memphis, USA. Each data sample is a single-timestamp winter snapshot (1980-2018), comprising $1245$ spatial grid points. At each point, four meteorological variables are recorded: temperature, pressure, the eastward wind component, and the northward wind component. The years are divided to 34K training samples and 17.5K evaluation samples. 
Random subsets of the total $1245$ points are drawn to form the context set and target set, and the task is to extrapolate the output (4D meteorological vector) at target points, given their input (2D coordinates) alongside the input-output pairs of context points. 
Baseline Models include CNP \cite{garnelo2018conditional}, NP \cite{garnelo2018neural}, ANP \cite{kimattentive} and TNP \cite{nguyen2022transformer}, details in Appendix \ref{app:era5}.

\paragraph{Results and Analysis.}
We report the target set log-likelihood, and the results are shown in Table \ref{tab:era5} and Figure \ref{fig:era5_sub}.  
Results show that fine-tuning yields consistent and rapid performance gains across various pre-training stages. 
Note that the chosen likelihood metric exhibits a very small standard deviation, indicating high confidence in the observed improvement.
The inclusion of GAV reduces model variance, while its ablation (replacement with a simple MLP) results in an increase.
These show the effectiveness of our method on complex datasets (meteorological data) and under complex architectures (Neural Process).

A comparison of Figure \ref{fig:combined_results} and their training errors (Figures~\ref{fig:kitchen_curve_loss}--\ref{fig:era5_curve_loss}) shows that MLE-based policy training is simpler than higher-dimensional prediction, performing well even with limited pre-training. Although robotic policies involve complex multimodal inputs, their low generated variable dimensionality (around a dozen) eases ODE simulation. By contrast, weather data spans hundreds of dimensions. As shown in the appendix, high‑dimensional image generation remains challenging for MLE, highlighting the need for more efficient training methods in such regimes.

\section{Conclusion}
This paper identifies some of inherent limitations of FM training and proposes a methodology that leverages MLE fine-tuning. 
We futher incorporate a GAV term in the fine-tuning framework to endow the flow with enhanced stability and robustnes.
Experimental results robustly demonstrate the efficacy of our fine-tuning strategy.
In future work, we will explore more stable and efficient MLE optimization methods, such as incorporating appropriate regularization, to ensure strong performance on high-dimensional generation tasks like images.
We will also examine the theoretical and experimental impact of more advanced implicit multi-step or multi-stage numerical methods on inference performance.
Furthermore, we incorporate more structure into GAV---for instance, by adding a velocity-augmentation term and applying friction-like negative feedback to velocity.


\section*{Impact Statement}

This paper presents work whose goal is to advance the field of Machine
Learning. There are many potential societal consequences of our work, none
which we feel must be specifically highlighted here.

\bibliography{example_paper}
\bibliographystyle{icml2026}

\newpage
\appendix
\onecolumn

\counterwithin{equation}{section}
\counterwithin{algorithm}{section}
\counterwithin{figure}{section}

\section*{\centering \LARGE Appendix}




\section{Proofs and Technical Details}\label{appSec:proof}

\subsection{Theorem \ref{th:errBound}}
We provide the proof of Theorem \ref{th:errBound} below.
\begin{proof}
	To be clear and concise, we define and rearrange some notations. 
	For an initial value $x_0$, let $x(t_n) := \psi_t(x_0)$ and $x_n := \hat{\phi}_n(x_0)$ represent the ground truth and the model output, respectively. When it does not cause ambiguity, we will omit $x_0$ on the right-hand side of the formula.  
    Denote $\delta(x_0)$ to be the upper bound: $\|v_{\theta}(t,\psi_t(x_0))-u_t(\psi_t(x_0)) \|_{\infty} \le \delta(x_0)$, then we have $\mathbb{E}_{p(x_0)} \delta(x_0) \le \delta$.
	Then we can derive that
	\begin{equation}\label{key:8.19-2}
		\begin{aligned}
			\varepsilon_{n+1}(x_0) &= x(t_{n+1}) - x_{n+1} \\
			&= x(t_n) + \int_{t_n}^{t_{n+1}} u_{\tau}(x(\tau)) \, d\tau - x_n - v_{t_n}(x_n) \\
			&= \varepsilon_{n-1} + \int_{t_n}^{t_{n+1}} u_{\tau}(x(\tau)) - v_{t_n}(x_n) \, d\tau \\
			= \varepsilon_{n-1} & + \int_{t_n}^{t_{n+1}} u_{\tau}(x(\tau)) - u_{t_n}(x(t_n)) + u_{t_n}(x(t_n)) -  u_{t_n}(x_n) + u_{t_n}(x_n) - v_{t_n}(x_n) \, d\tau .
		\end{aligned}
	\end{equation}
	We will split the integral in expression (\ref{key:8.19-2}) into three parts and estimate them separately. The first part
	\begin{equation}\label{key:8.19-3}
		\begin{aligned}
			& \ |\int_{t_n}^{t_{n+1}} u_{\tau}( x(\tau)) - u_{t_n}(x(t_n)) \, d\tau| \\
			&=| \int_{t_n}^{t_{n+1}} x'(\tau) - x'(t_n) \, d\tau| \\
			&=| \int_{t_n}^{t_{n+1}} x''(t_n + \theta (\tau - t_n)) (\tau - t_n) \, d\tau| \\
			&=| x''(t_n + \theta (\tau - t_n)) \int_{t_n}^{t_{n+1}} (\tau - t_n) \, d\tau| \\
			&=| \frac{1}{2} \tau_n^2 x''(t_n + \theta (\bar{t} - t_n))|\\
			&\le \frac{1}{2}M\tau_n^2 , 
		\end{aligned}
	\end{equation}
	where \( 0 < \theta < 1 \), \( \bar{t} \in (t_n, t_{n+1}) \), \( M = \max_{0 \leq t \leq 1} |x''(t)| \). The second and third equalities in (\ref{key:8.19-3}) use the Differentiation Mean Value Theorem (MVT) and Integration MVT, respectively.
	Using the Lipschitz condition, the second part in (\ref{key:8.19-2}) derives \[   |\int_{t_n}^{t_{n+1}} u_{t_n}(x(t_n)) - u_{t_n}(x_n) | =\tau_n\, |(u_{t_n}(x(t_n)) - u_{t_n}(x_n))| \le L_u\tau_n |x(t_n)-x_n|= L_u \tau_n \varepsilon_n  .  \]
	The third part \( | \int_{t_n}^{t_{n+1}}  u_{t_n}(x_n) - v_{t_n}(x_n) d \tau | \le \delta(x_0) \tau_n \). Plugging these into (\ref{key:8.19-2}) and taking the absolute values yield 
	\begin{equation}\label{key8.19-8}
		|\varepsilon_{n+1} (x_0)| \le (1+L_u \tau_n) |\varepsilon_{n}| + \delta(x_0) \tau_n + \frac{1}{2}M\tau_n^2,
	\end{equation}
	or equivalently,
	\begin{equation}\label{key:8.19-4}
		|\varepsilon_{n+1}|-|\varepsilon_{n}| \le L_u \tau_n |\varepsilon_{n}| + \delta(x_0) \tau_n + \frac{1}{2}M\tau_n^2.
	\end{equation}
	Summing $n$ in (\ref{key:8.19-4})   from $n = 0$ to $m-1$, we have
	\begin{equation*}
		|\varepsilon_{m}(x_0)| \le L_u \Sigma_{j=0}^{m-1} \tau_j |\varepsilon_{j}| + \Sigma_{j=0}^{m-1} (\delta(x_0) \tau_j + \frac{1}{2}M\tau_j^2) + |\varepsilon_{0}|.
	\end{equation*} 
	Appling Gr\"onwall inequality (Lemma \ref{Gronwall}) we get
	\begin{equation*}
		|\varepsilon_{m}(x_0)| \le \exp{(L_u t_{m-1})} \left( \Sigma_{j=0}^{m-1} (\delta(x_0) \tau_j + \frac{1}{2}M\tau_j^2) + |\varepsilon_{0}| \right)	.
	\end{equation*} 
	By setting $m=N$ and taking the expectation with respect to $x_0$ on both sides, we obtain the first estimate in Theorem \ref{th:errBound}.
	It is noted that Lemma 1 employs specialized scaling techniques designed for variable step-size schemes. 
	On a uniform grid, i.e., $\forall \,  i \in[0,N] $, $\tau_i=\tau_0$, we can have a shaper error estimate. It follows directly from (\ref{key8.19-8}) that 
	\begin{equation*}
		\begin{aligned}
			|\varepsilon_{n+1}(x_0)| &\leq (1 + \tau_0 L_u) |\varepsilon_n| + R \\
			&= (1 + \tau_0 L_u)^2 |\varepsilon_{n-1}| + (1 + \tau_0 L_u)R + R \\
			&\leq \cdots \\
			&\leq (1 + \tau_0 L_u)^{n+1} |\varepsilon_0| + \left[ (1 + \tau_0 L_u)^n + (1 + \tau_0 L_u)^{n-1} + \cdots + 1 \right] R(x_0) \, ,
		\end{aligned}
	\end{equation*}
	where $R:= \delta(x_0) \tau_0 + \frac{1}{2}M\tau_0^2$. 
	Therefore, 
	\begin{equation*}
		\begin{aligned}
			|\varepsilon_n(x_0)| &\leq (1 + \tau_0 L_u)^n |\varepsilon_0| + \left[ \sum_{j=0}^{n-1} (1 + \tau_0 L_u)^j \right] R(x_0) \\
			&\leq (1 + \tau_0 L_u)^n |\varepsilon_0| + \frac{R(x_0)}{\tau_0 L_u} \left[ (1 + \tau_0 L_u)^n - 1 \right].
		\end{aligned}
	\end{equation*}
	Considering $\exp{(n\tau_0 L_u)}>(1+\tau_0 L_u)^n$, we can obtain 
	\begin{equation}\label{key:8.19-6}
		\begin{aligned}
			|\varepsilon_n(x_0)| \leq \exp({L_u\tau_0 n})  |\varepsilon_0| + \frac{M\tau_0+2\delta(x_0)}{2L}(\exp{(L_u\tau_0 n})-1).
		\end{aligned}
	\end{equation}
	Letting $n=N$ and taking the expectation with respect to $x_0$ on both sides yield the second estimate in Theorem \ref{th:errBound}.
\end{proof}
\begin{remark}	
	The variable step-size method, while often more efficient, introduces a degree of uncontrollability due to its excessive degrees of freedom, thereby raising the upper bound.
\end{remark}
\begin{remark}
    This theorem is illustrated using the very basic explicit Euler method as an example. In subsequent work, we will further explore the performance of higher-order numerical methods in approximating the vector field of FM, such as Runge-Kutta (multi-stage) or BDF (multi-step) methods \cite{hairer1993solving,HairerWanner2002,li2022stability}.
\end{remark}
For the sake of completeness, in the following we put the standard discrete Gr\"onwall inequality, e.g., \citet[Lemma 3.1]{liao2021analysis}, and its proof.
\begin{lemma}\label{Gronwall}
	Let $\lambda \geq 0$, the time sequences $\{\xi_k\}_{k=0}^N$ and $\{V_k\}_{k=1}^N$ be nonnegative. If
	\[
	V_n \leq \lambda \sum_{j=1}^{n-1} \tau_j V_j + \sum_{j=0}^n \xi_j \quad \text{for } 1 \leq n \leq N,
	\]
	then it holds that
	\[
	V_n \leq \exp(\lambda t_{n-1}) \sum_{j=0}^n \xi_j \quad \text{for } 1 \leq n \leq N.
	\]
\end{lemma}

\begin{proof}
	Under the induction hypothesis $V_j \leq \exp(\lambda t_{j-1}) \sum_{k=0}^j \xi_k$ for $1 \leq j \leq n-1$, the desired inequality for the index $n$ follows directly from
	\[
	\lambda \sum_{j=1}^{n-1} \tau_j \exp(\lambda t_{j-1}) \leq \lambda \int_0^{t_{n-1}} \exp(\lambda t) \, dt = \exp(\lambda t_{n-1}) - 1.
	\]
	The principle of induction completes the proof.
\end{proof}

\subsection{Theorem \ref{th:mle}}

We present the proof of Theorem \ref{th:mle} in the following.
\begin{proof}
	Under the Assumption \ref{asmp:mle} , we have
	\begin{equation}\label{eq:9.3-1}
		\begin{aligned}
			\log p_1(x|x_0,x_1) &= \log \frac{\exp\left(-\frac{1}{2}(x-x_1)^T \Sigma^{-1}(x-x_1)\right)}{\sqrt{(2\pi)^d |\Sigma|}} \\ 
		\end{aligned}
	\end{equation}
	When $\sigma=\mathrm{diag}(\sigma^1,\cdots,\sigma^d)$ is a diagonal matrix with all positive entries, it follows from \ref{eq:9.3-1} that
	\begin{equation}\label{eq:9.3-2}
		\begin{aligned}
			\log p_1(x|x_0,x_1) = -\frac{1}{2} \sum_{i=1}^{d} \frac{(x^i - x_1^{i})^2}{\sigma^{i}} - \log \sqrt{(2\pi)^d \prod_{i=1}^{d} \sigma^{d}}
		\end{aligned}
	\end{equation}
	
	Note that the second term is independent of $x$. Moreover, considering the relation $\psi_1(x_0|x_1)=x_1$ and $\varepsilon_N(x_0|x_1)=\psi_1(x_0|x_1)-\hat{\phi}_N(x_0)$,
	maximizing
	$\mathbb{E}_{q(x_0, x_1)} \left[\log  p_{1}\left(\hat{\phi}_N(x_0)|x_1\right) \right]$
	is equivalent to minimizing
	\(
	\mathbb{E}_{q(x_0, x_1)} \left[
	\frac{1}{2} \sum_{i=1}^{d} (\varepsilon_N^i(x_0|x_1))^2/\sigma^{i} \right]
	\).
	
	In the case where $\Sigma$ is a scalar matrix, i.e., $\Sigma=\sigma I$ with $\sigma \in \mathbb{R}_{+}$, following the same procedure and discarding the irrelevant coefficient leads to the loss function (\ref{eq:mle-loss}).
\end{proof}

\begin{remark}[Gaussian Assumption]
    According to the maximum entropy principle, among all possible distributions that satisfy the known constraints, the one with the maximum entropy is the inference that introduces the fewest additional assumptions and is the most impartial \cite{jaynes1957information}. Given constraints on only the mean and covariance, this distribution is precisely Gaussian \cite{jaynes1982rationale}. 
    Even when viewed as a kernel method, the choice of a Gaussian kernel is justified as it provides a consistent estimator of the underlying distribution \cite{wied2012consistency}.
    In fact, this posterior Assumption \ref{asmp:mle}---a multivariate Gaussian with diagonal covariance---aligns with those made in many classical methods, including the Variational Autoencoder (VAE) \cite{kingma2013auto}.
    Similarly, this is just a (simplifying) choice, and not a limitation of our method.
\end{remark}

\subsection{Theorem \ref{thm:iss-con}}\label{app:iss-con}
The theoretical foundation of Theorem \ref{thm:iss-con} is inspired by recent advances in dynamic system modeling, specifically works such as ControlSynth Neural ODE \citep{mei2024controlsynth} and analyses of generalized Persidskii systems \citep{efimov2021analysis,mei2022convergence}. 
We apply these principles to generative models characterized by ODEs.
\subsubsection{Notations and Definitions}\label{app:N_D_iss-con}
The symbol $\mathbb{R}$ represents
the set of real numbers, $\mathbb{R}_{+}=\left\{ \ell \in\mathbb{R}: \ell \geq0\right\} $, and $\mathbb{R}^{n}$ denotes the vector space of $n$-tuple
of real numbers. 
The transpose of a matrix $A\in\mathbb{R}^{n\times n}$ is denoted by $A^{\top}$. Let $I$ stand for the identity matrix. The symbol $\rVert\cdot\rVert$ refers to the Euclidean norm on $\mathbb{R}^{n}$.

For a Lebesgue measurable function $u\colon\mathbb{R}\rightarrow\mathbb{R}^{q}$,
define the norm $\rVert u\rVert_{(t_{1},t_{2})}=\text{ess}\sup_{t\in(t_{1},t_{2})}\rVert u(t)\rVert$
for $(t_{1},t_{2})\subseteq\mathbb{R}$. We denote by $\mathscr{L}_{\infty}^{q}$
the space of functions $u$ with $\rVert u\rVert_{\infty}:=\rVert u\rVert_{(-\infty,+\infty)}<+\infty$.

A continuous function $\alpha:\mathbb{R}_{+}\to\mathbb{R}_{+}$ belongs
to class $\mathscr{K}$ if it is strictly increasing and $\alpha(0)=0$, and $\mathscr{K}_{\infty}$ means that $\alpha$ is also unbounded. 
A continuous function $\beta:\mathbb{R}_{+}\times\mathbb{R}_{+}\to\mathbb{R}_{+}$
belongs to class $\mathscr{KL}$ if $\beta(\cdot,r)\in\mathscr{K}$
and $\beta(r,\cdot)$ is a decreasing to zero function for any fixed
$r>0$. 

The definitions of ISS and contraction are provided below.
\begin{definition}
	A forward complete system (\ref{eq:sys_delta}) is
	input-to-state stable (\underline{ISS}) if there exist $\beta \in \mathscr{KL}$ and $\gamma \in \mathscr{KL}$ such that
	\[
	\|x(t,x_0,u)\| \leq \beta (\|x_0\|,t) + \gamma(\|u\|_{\infty}), \quad \forall t \in \mathbb{R}_{+},
	\]
	for any $x_0 \in \mathbb{R}^n$ and $u \in \mathcal{L}_{\infty}^q$.
    If nearby trajectories converge to one another in a region of the state space, then the model is \underline{contracting} in this contraction region \cite{lohmiller1998contraction}.
\end{definition}
\begin{remark}
    The model (\ref{eq:sys_delta}) is \underline{convergent} if it admits a unique bounded solution for $t \in \mathbb{R}$ that is globally asymptotically stable (GAS). Convergence is a global property, whereas contraction is only local within each contraction region.
\end{remark}

ISS refers to the property that the model output remains stable and bounded in the presence of external inputs, thereby avoiding divergence behavior as illustrated by the blue curves in Figure \ref{fig:flow_c}. 
ISS property is particularly crucial in fields such as robotic manipulation. We obviously do not want minor disturbances to be excessively amplified, causing severe jitter in movements.
Furthermore, the contraction property indicates that the model exhibits robustness against small disturbance noises, as shown by the green curves in Figure \ref{fig:flow_c} and Figure \ref{fig:flow_con_2dim}. 
Each contraction region can correspond to a specific semantic meaning in the latent space. 
We may even achieve single-step inference without relying on distillation and without loss of accuracy by directly mapping each contraction region to its corresponding equilibrium point. 

\subsubsection{Further details of system (\ref{eq:sys_delta})}
Substituting the expression for the viscity $\delta(\cdot)$, we obtain the complete form of the system
\begin{equation} \label{eq:controlsynth}
	\dot{x}(t) = A_0 x(t) + \sum_{j=1}^M A_j f_j(W_j x(t)) +  \tilde{v}( \bm o(t)),
\end{equation}
where $x_t := x(t) \in \mathbb{R}^{n}$ is the state vector; $A_{\cdot},W_{\cdot}$ are weight matrices; the external inputs or conditions $\bm o_t$.
$\bm o_t := \bm o(t) \in \bm O \subset \mathbb{R}^m$, $\bm o \in \mathscr{L}_{\infty}^{m}$; $f_j = [f_j^1 \dots f_j^{k_j}]^\top$ $\left(f_j: \mathbb{R}^{k_j} \to \mathbb{R}^{k_j} \right)$ and $\tilde{v}: \bm O \to  \mathbb{R}^n$ 
ensuring the existence of the solutions of the neural network (NN)~(\ref{eq:sys_delta}) 
at least locally in time, and $g = [g_1 \dots g _n]^\top$;  \emph{w.l.o.g.}, the time $t$ is set as $t \geq 0$. 
The system (\ref{eq:sys_delta}) is forward complete, i.e., for all $x_0 \in \mathbb{R}^n$ and $\bm o \in \mathcal{L}_{\infty}^q$, the solution $x(t,x_0,\bm o)$ is uniquely defined for all $t \in \mathbb{R}_+$.

\subsubsection{Details and Proof of the Theorem}

Theorem \ref{thm:iss-con} can be decomposed into two theorems: the first concerns ISS, while the second further advances to contraction. 
We begin by introducing the following two assumptions regarding the nonlieanr function $f$.
\begin{assumption}\label{asmp:sign}
	For any $i \in \{1, \ldots, k_j\}$ and $j \in \{1, \ldots, M\}$, $s f_j^i(s) > 0$ for all $s \in \mathbb{R}\backslash\{0\}$.
\end{assumption}
\begin{assumption}\label{asmp:con-inc}
	Functions $f_j^i$ are continuous and strictly increasing for any $i \in \{1, \ldots, k_j\}$ and $j \in \{1, \ldots, M\}$.
\end{assumption}
Assumption \ref{asmp:sign} applies to many activation functions, such as $\tanh$ and parametric ReLU. 
With a reordering of nonlinearities and their decomposition, there exists an index $\omega \in \{0, \ldots, M\}$ such that for all $1 \le s \le \omega$ and $1 \le i \le k_s$, $\lim\limits_{\nu \to \pm \infty} f_s^i(\nu) = \pm \infty$. Also, there exists $\zeta \in \{\omega, \ldots, M\}$ such that for all $1 \le s \le  \zeta$, $1 \le i \le k_s$, we have $\lim\limits_{\nu \to \pm \infty} \int\limits_0^\nu f_s^i(r) dr = +\infty$.

Now we can introduce the ISS theorem. 
\begin{theorem}\label{th:iss}
	Let Assumptions \ref{asmp:sign}-\ref{asmp:con-inc} be satisfied. If there exist positive semidefinite symmetric matrices $P$; positive semidefinite diagonal matrices $\{ \Lambda^i = \text{diag}(\Lambda_1^i, \ldots, \Lambda_n^i) \}_{i=1}^M$, $\{ \Xi^s \}_{s=0}^M$, $\{ \Upsilon_{s,r} \}_{0 \leq s < r \leq M}$; positive definite symmetric matrix $\Phi$ such that the following linear matrix inequalities hold true:
	\begin{equation}\label{eq:ISS-condition}
		P + \sum_{j=1}^{\zeta} \Lambda^j > 0; \quad Q = Q^T \leq 0; \quad  \sum_{j=1}^M \Upsilon_{0,j} + \sum_{s=1}^{\omega} \Xi^s + \sum_{s=1}^{\omega} \sum_{r=s+1}^{\omega} \Upsilon_{s,r} > 0.
	\end{equation} 
	where
	\begin{align*}
		Q_{1,1} &= A_0^\top P + P A_0 + \Xi^0; \quad Q_{j+1,j+1} = A_j^\top W_j^\top \Lambda^j + \Lambda^j W_j A_j + \Xi^j; \\
		Q_{1,j+1} &= P A_j + A_0^\top W_j^\top \Lambda^j + W_j^\top \Upsilon_{0,j}; \ Q_{s+1,r+1} = A_s^\top W_r^\top \Lambda^r + \Lambda^s W_s A_r + W_s^\top W_r \Upsilon_{s,r} W_r^\top W_s; \\
		Q_{1,M+2} &= P; \quad Q_{M+2,M+2} = -\Phi; \quad Q_{j+1,M+2} = \Lambda^j W_j.
	\end{align*}
	then system (\ref{eq:sys_delta}) is ISS. 	
\end{theorem}

Next, to analyse contraction, we consider another trajectory of the model (\ref{eq:controlsynth})
$
\dot{y}(t) = \sum_{j=1}^M A_j f_j(W_j y(t)) + \tilde{v}( \bm o(t))
$
with the same input but different initial conditions $y(0) \in \mathbb{R}^n$. Let $\xi := y - x$. Then the corresponding error system is
\begin{equation}\label{key:8.22-1}
	\dot{\xi} = A_0 p_j(\xi) + \sum_{j=1}^M A_j p_j(x, \xi),
\end{equation}
where $p_j(x, \xi) = f_j(W_j(\xi + x)) - f_j(W_j x)$. Note that for any fixed $x \in \mathbb{R}^n$, the functions $p_j$ in the variable $\xi \in \mathbb{R}^n$ satisfy the properties in Assumptions \ref{asmp:sign}-\ref{asmp:con-inc} (with a different Lipschitz constant).
\begin{theorem}\label{th:contraction}
	Let Assumptions \ref{asmp:sign}-\ref{asmp:con-inc} and conditions in Theorem \ref{th:iss} be satisfied, 
	and in the bounded domain (determined by the ISS property), the functions $f_j^i$ are Lipschitz continuous with Lipschitz constants $L_j^i$.
	If there exist positive semidefinite symmetric matrices $\tilde{P}$; positive semidefinite diagonal matrices $\{ \tilde{\Lambda}^i = \text{diag}(\tilde{\Lambda}_1^i, \ldots, \tilde{\Lambda}_n^i) \}_{i=1}^M$, $\{ \tilde{\Upsilon}_{j,r} \}_{j,r=1}^M$, $\{ \Gamma_j \}_{j=1}^M$, $\{ \Omega_j \}_{j=1}^M$; positive definite symmetric matrix $\Phi$; and positive scalars $\gamma, \theta$ such that the following linear matrix inequalities hold true:
	\begin{equation}\label{eq:con-condition}
		\begin{aligned}	
			\quad \tilde{Q} = \tilde{Q}^T \leq 0; \quad  \Gamma_j - \gamma L^j \geq 0; \quad 	\Omega_j - \theta L^j \geq 0; \\
			\sum_{j=1}^M \left( \Gamma_j - \gamma L^j  + \Omega_j -  \theta L^j \right) + \sum_{j=1}^{M} \sum_{r=1}^{M} \tilde{\Upsilon}_{j,r} >0, 
		\end{aligned}
	\end{equation}
	where
	\begin{align*}
		\tilde{Q}_{1,1} &= A_0^\top \tilde{P} + P A_0 + \tilde{\Xi}^0; \quad \tilde{Q}_{2,2} = -2\gamma I; \quad \tilde{Q}_{1,2} = P A + \Gamma; \quad \tilde{Q}_{1,3} = A_0^\top \Delta + \Omega; \\
		\tilde{Q}_{2,3} &= A^\top \Delta + \tilde{\Upsilon}; \quad \tilde{Q}_{3,3} = -2\theta I; \quad A = \left[ \begin{array}{ccc} A_1 & \cdots & A_M \end{array} \right]; \quad \Gamma = \left[ \begin{array}{ccc} W_1^\top \Gamma_1 & \cdots & W_M^\top \Gamma_M \end{array} \right]; \\
		\Delta =& \left[ \begin{array}{ccc} W_1^\top \Lambda^1 & \cdots & W_M^\top \Lambda^M \end{array} \right]; \  
		\Omega = \left[ \begin{array}{ccc} W_1^\top \Omega_1 & \cdots & W_M^\top \Omega_M \end{array} \right]; \  
		\tilde{\Upsilon} = (W_j^\top W_j \tilde{\Upsilon}_{j,r} W_r^\top W_r)_{j,r=1}^M,
	\end{align*}
	then system (\ref{eq:sys_delta}) (with trajectory $x$) is contracting. 
	If we define $\tilde{V}(\zeta) = \zeta^\top \tilde{P} \zeta + 2 \sum_{j = 1}^M \sum_{i=1}^{k_j} \tilde{\Lambda}^j_i \int_{0}^{W_j^i \zeta}  f_j^i(s) ds$, then the contraction region of $x_0$ contains
	$\{x_0+\xi|V(\xi)\le \max_{\zeta \in \mathbb{R}^d } \tilde{V}(\zeta) \}$.
\end{theorem}

We provide the proofs of the two theorems below. First, we present Theorem \ref{th:iss}. 

\begin{proof}[Proof of Theorem \ref{th:iss}]
Similar proofs can be found in \citet{mei2024controlsynth, mei2022convergence, efimov2021analysis}. Here, we briefly outline the general process.

	Consider a Lyapunov function 
	\begin{equation*}
		V(x) = x^\top P x + 2 \sum_{j = 1}^M \sum_{i=1}^{k_j} \Lambda^j_i \int_{0}^{W_j^i x}  f_j^i(s) ds,   
	\end{equation*}
	where the vector $W_j^i$ is the $i$-th row of the matrix $W_j$. It is positive definite and radially unbounded due to Finsler’s Lemma under the condition~\eqref{eq:ISS-condition} and Assumption \ref{asmp:sign}. Then, taking the derivative of $V(x)$, one has  
	\begin{eqnarray*}
		\dot{V} & = & \begin{bmatrix}
			x\\
			f_1(W_{1}x)\\
			\vdots\\
			f_M(W_{M}x)\\
			\tilde{v}( \bm o)
		\end{bmatrix}^{\top}Q\begin{bmatrix}
			x\\
			f_1(W_{1}x)\\
			\vdots\\
			f_M(W_{M}x)\\
			\tilde{v}( \bm o)
		\end{bmatrix}-x^{\top}\Xi^{0}x \\  & &
		-\sum_{j=1}^{M}f_j(W_{j}x)^{\top}\Xi^{j}f_j(W_{j}x) -2\sum_{j=1}^{M}x^{\top}W_{j}^{\top}\Upsilon_{0,j}f_j(W_{j}x)
		\\
		& & -2\sum_{s=1}^{M-1}\sum_{r=s+1}^{M}f_s(W_{s}x)^{\top} {W_{s}^\top} W_{s}\Upsilon_{s,r}W_{r}^{\top} {W_{r}} f_r(W_{r}x)+\tilde{v}( \bm o)^{\top}\Phi \tilde{v}( \bm o)\\ &
		\leq &
		-x^{\top}\Xi^{0}x-\sum_{j=1}^{M}f_j(W_{j}x)^{\top}\Xi^{j}f_j(W_{j}x)
		-2\sum_{j=1}^{M}x^{\top}W_{j}^{\top}\Upsilon_{0,j}f_j(W_{j}x)\\ &
		&
		-2\sum_{s=1}^{M-1}\sum_{r=s+1}^{M}f_s(W_{s}x)^{\top} {W_{s}^\top} W_{s}\Upsilon_{s,r}W_{r}^{\top} {W_{r}} f_r(W_{r}x)+\tilde{v}( \bm o)^{\top}\Phi \tilde{v}( \bm o)\\
		&
		\leq & -\alpha(V) +\tilde{v}( \bm o)^{\top}\Phi \tilde{v}( \bm o),
	\end{eqnarray*}
	for a function $\alpha \in \mathscr{K}_{\infty}$. 
	Under \citet[Theorem 1]{sontag1995characterizations},
	we can verify the first condition of the ISS property due to the form of $V$, and the second relation can be recovered via $V\geq\alpha^{-1} \left(2\tilde{v}( \bm o) ^{\top}\Phi \tilde{v}( \bm o) \right)\Rightarrow\dot{V}\leq-\frac{1}{2}\alpha(V)$. This means that the ISS property of the NN~\eqref{eq:controlsynth} is guaranteed, and so is the boundedness of its solution.
\end{proof}

Next, we now proceed to prove the contraction Theorem \ref{th:contraction}.
To analyze the contraction property of (\ref{key:8.22-1}), first we need the following lemma.
\begin{lemma}\label{len:con}
	Under Assumption \ref{asmp:con-inc}, we have 
	$
	p_j(x,\xi)^\top p_j(x,\xi) \leq   \xi^\top W_j^\top L^j p_j(x,\xi) 
	$
	and
	$
	f_j(W_j \xi)^\top f_j(W_j \xi) \leq   \xi^\top W_j^\top L^j f_j(W_j \xi).
	$
\end{lemma}
\begin{proof}
	It follows from the Lipschitz continuity Assumption \ref{asmp:con-inc} that $|p_j^i(x,\xi)|=|f_j^i( (W_jx)^i + (W_j \xi)^i) - f_j^i((W_j x)^i)| \le L_j^i |(W_j \xi)^i| $. Here, the superscript $i$ denotes the $i$-th component of the vector.
	When $(W_j x)^i\ge0$, we have $p_j^i(x,\xi)=f_j^i( (W_jx)^i + (W_j \xi)^i) - f_j^i((W_j x)^i)\ge0$ due to the Monotonicity in Assumption \ref{asmp:con-inc}. Then
	$p_j^i(x,\xi)\le L_j^i (W_j \xi)^i$. 
	Multiplying both sides by a non-negative number $p_j^i(x,\xi)$, we get
	$p_j^i(x,\xi)^2  \le L_j^i p_j^i(x,\xi) (W_j x)^i$. 
	When $(W_j x)^i\le0$, we have $p_j^i(x,\xi)=f_j^i( (W_jx)^i + (W_j \xi)^i) - f_j^i((W_j x)^i)\le0$ due to the same Monotonicity Assumption, leading to
	$-p_j^i(x,\xi)\le -L_j^i (W_j \xi)^i$. 
	Multiplying both sides by a non-negative number $-p_j^i(x,\xi)$, we get the same result $p_j^i(x,\xi)^2  \le L_j^i p_j^i(x,\xi) (W_j x)^i$. 
	
	Summing over $i$ on both sides gives $\sum_i p_j^i(x,\xi)^2 \le \sum_i L_j^i p_j^i(x,\xi) (W_j x)^i$. Or equivalently, in a compact from, $p_j(x,\xi)^\top p_j(x,\xi)\le (W_j x)^\top L_j p_j(x,\xi)= x^\top W_j^\top L_j p_j(x,\xi) $. This completes the proof of the first part of the lemma. 
	By noting the relation of $f(W_jx)=f(W_jx-0)=p_j(0,\xi)$, the second part of the lemma holds naturally.
\end{proof}
We now begin the proof of the theorem.
\begin{proof}[Proof of Theorem \ref{th:contraction}]
	Consider an positive definite function   
	\begin{equation*}
		\tilde{V}(\xi) = \xi^\top \tilde{P} \xi + 2 \sum_{j = 1}^M \sum_{i=1}^{k_j} \tilde{\Lambda}^j_i \int_{0}^{W_j^i \xi}  f_j^i(s) ds.   
	\end{equation*}
	Taking the time derivative of $\tilde{V}$:
	\begin{eqnarray*}
		\dot{\tilde{V}} & = & \left[\begin{array}{c}
			\xi \\
			p_{1}(x,\xi )\\
			\vdots\\
			p_{M}(x,\xi )\\
			f_{1}(W_1 \xi )\\
			\vdots\\
			f_{M}(W_M  \xi )
		\end{array}\right]^{\top}\tilde{Q}\left[\begin{array}{c}
			\xi \\
			p_{1}(x,\xi )\\
			\vdots\\
			p_{M}(x,\xi )\\
			f_{1}(W_1 \xi )\\
			\vdots\\
			f_{M}(W_M \xi )
		\end{array}\right] \\
		& & +\gamma\sum_{j=1}^{M}p_j(x,\xi )^{\top}p_j(x,\xi )+\theta\sum_{j=1}^{M}f_{j}^{\top}(W_j \xi )f_{j}(W_j \xi )\\
		&  & -\xi ^{\top}\tilde{\Xi}^{0}\xi -2\sum_{j=1}^{M}\xi ^{\top} W_j^\top \Gamma_{j}p_j(x,\xi )  -2\sum_{j=1}^{M}\xi ^{\top} W_j^\top \Omega_{j}f_{j}(W_j \xi )\\ 
		& & -2\sum_{j=1}^{M}\sum_{r=1}^{M} p_j(x,\xi )^{\top} W_j^\top W_j \tilde{\Upsilon}_{j,r} W_r^\top W_r f_{r}(W_r \xi). 
	\end{eqnarray*}
	Then, under Lemma~\ref{len:con}, it can be deduced that 
	\begin{eqnarray*}
		\dot{\tilde{V}} & \leq & 2\gamma\sum_{j=1}^{M}p_j(x,\xi )^{\top}p_j(x,\xi ) + 2\theta\sum_{j=1}^{M}f_{j}^{\top}(W_j \xi )f_{j}(W_j \xi )-\xi ^{\top}\tilde{\Xi}^{0}\xi -2\sum_{j=1}^{M}\xi ^{\top} W_j^\top \Gamma_{j}p_j(x,\xi ) \\ & &  - 2\sum_{j=1}^{M}\xi ^{\top} W_j^\top \Omega_{j}f_{j}(W_j \xi )-2\sum_{j=1}^{M}\sum_{r=1}^{M} p_j(x,\xi )^{\top} W_j^\top W_j \tilde{\Upsilon}_{j,r} W_r^\top W_r f_{r}(W_r \xi) \\ 
		& \leq &  -\xi^{\top}\tilde{\Xi}^{0} \xi\\
		&  & -2\sum_{j=1}^{M} \xi^{\top}  W_j^\top \left(  \Gamma_{j}-\gamma L^{j}\right) p_{j}(x,\xi)\\
		&  & -2\sum_{j=1}^{M} \xi^{\top} W_j^\top \left(\Omega_{j}-\theta L^{j}\right)f_{j}(W_j \xi)\\
		& &-2\sum_{j=1}^{M}p_{j}(x,\xi)^{\top} W_j^\top W_j \sum_{r=1}^{M}\left(\tilde{\Upsilon}_{j,r}\right) W_r^\top W_r 
		f_{ r}(W_r \xi).
	\end{eqnarray*}
	Therefore, with the conditions~\eqref{eq:con-condition}, we can substantiate that the error dynamics of system \eqref{eq:controlsynth} asymptotically approaches zero, meaning that the solution is contracting. 
	Moreover, by \citet[Theorem 4.9]{khalil2002nonlinear}. 
	the stability area of $\xi$ contains 
	$\{\xi|V(\xi)\le \max_{\zeta \in \mathbb{R}^d } \tilde{V}(\zeta) \}$. Since $y(t)=x(t)+\xi(t)$, we can determine that $\forall y_0 \in \{x_0+ \xi|V(\xi)\le \max_{\zeta \in \mathbb{R}^d } \tilde{V}(\zeta) \}$, $y(t) \to x(t)$.
	This completes the proof. 
\end{proof}

\begin{remark}
   Theorems \ref{th:iss} and \ref{th:contraction} establish the ISS and contraction property for systems of the form (\ref{eq:controlsynth}) via tractable linear inequalities.
For the contraction analysis of these more general forms, we can refer to \citet{lohmiller1998contraction} and \citet{li2025icode}, albeit at the expense of increased condition complexity. 
\end{remark}

\subsection{Corollary \ref{thm:iss-con} and the Direct Parameterization Scheme}\label{app:iss-con-cor}
We first provide the proof of Corollary \ref{thm:iss-con}.
\begin{proof}[Proof of Corollary \ref{thm:iss-con}]
The proof is straightforward. After a simple transformation, $\delta(x)=A_0 x(t) + \sum_{j=1}^M A_j f_j(D_j x(t))=A_0 x(t) + \sum_{j=1}^M A_j \hat{f}_j( x(t))$, where $\hat{f}_j^i=D_j^{(ii)}f_j^i(x^i(t))$. Given that all entries of the diagonal matrix $D_j$ are positive, it is evident that $\hat{f}$ also satisfies Assumptions \ref{asmp:sign}-\ref{asmp:con-inc}. The proof follows directly from the results of Mei et~al. \yrcite{mei2022convergence}.
\end{proof}

\paragraph{Direct Parameterization.}
Next, we detail how to leverage Corollary \ref{thm:iss-con} to construct a direct parameterization scheme. Note that we can construct a Hurwitz matrix via the Schur decomposition. Assuming all eigenvalues of $A_j$ are real, since we don't seem to have the need to use complex eigenvalues to characterize oscillations in generative models. To be more specific, we let $A_j=Q_jT_jQ_j^T$, where $T$ is an upper triangular matrix with negative diagonal elements, and $Q_j$ is an orthogonal matrix. We further use the Cayley transform to parameterize $Q_j$ as $Q_j=(I+K_j)(I-K_j)^{-1}$, where $I$ is the identity matrix and $K_j$ is an arbitrary skew-symmetric matrix ($K_j^T=-K_j$), which has $d(d-1)/2$ degrees of freedom. 
The computational complexity of matrix inversion in the Cayley transform can be addressed by using convolutions in the Fourier domain \cite{trockmanorthogonalizing}.
However, here we adopt a simpler and more straightforward strategy that relies on a low-rank assumption on $K_j$. Let $K_j=ef^T-fe^T$, where $e,f\in \mathbb{R}^d$. Then  using the Sherman-Morrison-Woodbury (SMW) formula, we have $(I-K)^{-1}=(I-ECF^T)^{-1}$, where $E=[e,f]\in \mathbb{R}^{d \times 2}, F=[f,e]\in \mathbb{R}^{d \times 2}, C=[0,-1;1,0]\in \mathbb{R}^{2 \times 2}$. Note that here we only require the inversion of a two-dimensional matrix.
The previously involved positive quantities can be obtained using $\exp(\cdot)$ or $\operatorname{softplus}(\cdot)$ functions. 
Thus, we have implemented a direct parameterization scheme that inherently satisfies conditions in Theorem \ref{thm:iss-con} during training.

\subsection{Algorithm Workflow}
Algorithm workflows of FM, MLE finetuning and residual finetuning with generalized artificial viscosity are presented in Algorithms \ref{alg:flow_train}-\ref{alg:GAV}.

\renewcommand{\algorithmicrequire}{\textbf{Input:}}
\renewcommand{\algorithmicensure}{\textbf{Output:}}

\noindent 
\begin{minipage}[t]{0.48\textwidth} 
\begin{algorithm}[H] 
\caption{Flow Matching Training}\label{alg:flow_train}
\begin{algorithmic}[1]
\REQUIRE observation $\bm O$, samples $\bm X_{1}$, source random waypoints $p_0$
\ENSURE flow $\mathbf{v}_{\bm \theta}$
\WHILE{not converged}
\STATE $\bm x_0 \sim p_0$; $\bm x_{1},\bm o \sim \bm X_{1},\bm O $, Sample a batch
\STATE (Optional) reassign the pairing between $\bm x_0$ and $\bm x_1$
\STATE $t \sim \mathcal{U}[0, 1]$, sample time steps
\STATE $\bm x_t = t\bm x_1+(1-t)\bm x_0$, linear interpolation
\STATE $\nabla_{\bm \theta} \left\| v_{\bm \theta}(\bm x_t, t | \bm{o}) - \underbrace{ \dot{\bm x}_t}_{ x_1-x_0} \right\|$, gradient step
\ENDWHILE
\STATE Stopping criteria: training epochs reached
\end{algorithmic}
\end{algorithm}
\end{minipage}
\hfill 
\begin{minipage}[t]{0.48\textwidth}
\begin{algorithm}[H]
\caption{MLE Finetuning}\label{alg:flow_finetune}
\begin{algorithmic}[1]
\REQUIRE observation $\bm O$, samples $\bm X_{1}$, source random waypoints $p_0$
\ENSURE flow $\mathbf{v}_{\bm \theta}$
\WHILE{not converged}
\STATE $\bm x_0 \sim p_0$; $\bm x_{1},\bm o \sim \bm X_{1},\bm O $,  Sample a batch
\STATE (Optional) reassign the pairing between $\bm x_0$ and $\bm x_1$
\STATE Numerical Solve $\frac{d}{dt}\phi_t=  v_{\bm \theta}(\bm \phi_t, t | \bm{o})$ with $\phi_0=x_0$, get $\hat{\phi}_{t=1}$.
\STATE $\nabla_{\bm \theta} \left\| \hat{\phi}_{t=1} - x_1 \right\|$, gradient step
\ENDWHILE
\STATE Stopping criteria: training epochs reached
\end{algorithmic}
\end{algorithm}
\end{minipage}

\renewcommand{\algorithmicrequire}{\textbf{Input:}}
\renewcommand{\algorithmicensure}{\textbf{Output:}}
\begin{algorithm}[t]
\caption{MLE Finetuning with GAV}\label{alg:GAV}
\begin{algorithmic}[1]
\REQUIRE Condtion $\bm O$, samples $\bm X_{1}$, source random waypoints $p_0$.
Hyperparamters: a positive integer $M$, nonlinear functions $f_1,\cdots,f_M$, nonnegative activation function $\sigma(\cdot)$ (e.g., $\operatorname{exp}(\cdot)$).
Initial weights: $\tilde{\theta}$, $\tilde{D}_j,e_j,f_j\in \mathbb{R}^d$, $\tilde{T_j}\in \mathbb{R}^{d(d-1)/2}$, 
\ENSURE flow $\mathbf{v}_{\bm \theta}$
\STATE $D=\sigma(\tilde{D})$, $T_{ll}=-\sigma(\tilde{T}_{(l+1)l/2})$, $T_{lm}=\tilde{T}_{(l+1)l/2+m}$ for $l=1,\cdots,d$,  
and $m=m+l,\cdots,d$, $K_j=e_j f_j^T-f_j e_j^T $, 
\STATE $\text{Inv}(I-K_j)=(I-E_jCF_j^T)^{-1}$, where $E_j=[e_j,f_j]\in \mathbb{R}^{d \times 2}, F_j=[f_j,e_j]\in \mathbb{R}^{d \times 2}, C=[0,-1;1,0]\in \mathbb{R}^{2 \times 2}$, SMW formula
\STATE $Q_j=(I+K_j)\, \text{Inv}(I-K_j)$, Cayley transform
\STATE $A_j=Q_jT_jQ_j^T$, Schur decomposition
\STATE $\delta(x)=A_0 x + \sum_{j=1}^M A_j f_j(D_j x)$
\WHILE{not converged}
\STATE $\bm x_0 \sim p_0$; $\bm x_{1},\bm o \sim \bm X_{1},\bm O $, Sample a batch
\STATE (Optional) reassign the pairing between $\bm x_0$ and $\bm x_1$
\STATE Numerical Solve $\frac{d}{dt} \phi_t = v_t(\phi_t; \theta)$ with initial value $\phi_0 = x_0$ for $t\in[0,1]$, get $\hat{\phi}_{t=1}(x_0)$
\STATE Numerical Solve $\frac{d}{dt} \phi_t  = \tilde{v}_t(\bm o; \tilde{\theta})+\delta (\phi_t(x))$ with initial value $\phi_1 = \hat{\phi}_{t=1}(x_0)$ for $t\in[1,2]$, get $\hat{\phi}_{t=2}(x_0)$

\STATE Compute Loss $ \left\| \hat{\phi}_{t=2}(x_0) -x_1 \right\|$, gradient step
\ENDWHILE
\STATE Stopping criteria: training epochs reached
\end{algorithmic}
\end{algorithm}

\section{Mathematical Analysis of Viscous Term as Negative Feedback}\label{app:viscous}
\subsection*{From PDE to ODE System via Spatial Discretization}
For simplicity, we take the one-dimensional viscous Burgers equation with periodic boundary conditions as an example:
\[
\frac{\partial u}{\partial t} + u\frac{\partial u}{\partial x} = \nu \frac{\partial^2 u}{\partial x^2}, \quad x \in [0, L], \quad \nu > 0
\].

Discretizing the domain into $N$ equally spaced points with spacing $\Delta x = L/N$, we define the state vector $\mathbf{u}(t) = [u_0(t), u_1(t), \dots, u_{N-1}(t)]^T \in \mathbb{R}^N$. Using central differences for both the convective and viscous terms, we obtain the semi-discrete system:
\[
\frac{d\mathbf{u}}{dt} = \mathbf{f}(\mathbf{u}) + \nu \mathbf{A} \mathbf{u}
\]
where $\mathbf{f}(\mathbf{u})$ represents the discretized nonlinear convection and $\nu \mathbf{A} \mathbf{u}$ is the discretized viscous term. The matrix $\mathbf{A}$ has the circulant tridiagonal structure:
\[
\mathbf{A} = \frac{1}{(\Delta x)^2}
\begin{bmatrix}
-2 & 1 & 0 & \cdots & 0 & 1 \\
1 & -2 & 1 & \cdots & 0 & 0 \\
0 & 1 & -2 & \cdots & 0 & 0 \\
\vdots & \vdots & \vdots & \ddots & \vdots & \vdots \\
1 & 0 & 0 & \cdots & 1 & -2
\end{bmatrix}
\]

\subsection*{Eigenvalue Analysis Revealing Negative Feedback Structure}
The eigenvalues of $\mathbf{A}$ can be computed analytically due to its circulant structure:
\[
\lambda_k(\mathbf{A}) = -\frac{4}{(\Delta x)^2} \sin^2\left(\frac{\pi k}{N}\right), \quad k = 0, 1, \dots, N-1
\]
\textbf{\textit{Crucially, all eigenvalues have non-positive real parts}}:
\begin{itemize}
    \item $\lambda_0 = 0$ corresponds to the constant mode (due to mass conservation with periodic boundaries)
    \item $\lambda_k < 0$ for $k = 1, 2, \dots, N-1$, with $\lambda_{N/2} = -4/(\Delta x)^2$ being the most negative for even $N$
\end{itemize}
This eigenvalue structure has profound implications. In Fourier space, the system decouples into modal equations:
\[
\frac{d\hat{u}_k}{dt} = \hat{f}_k(\hat{\mathbf{u}}) - \underbrace{|\nu \lambda_k|}_{\text{feedback gain}} \hat{u}_k, \quad k = 1,\dots,N-1
\]
The viscous term appears explicitly as linear negative feedback with gain $g_k = |\nu \lambda_k| = \frac{4\nu}{(\Delta x)^2} \sin^2(\pi k/N)$. The gain increases with wavenumber $k$, meaning higher frequency modes experience stronger damping---a characteristic feature of viscous dissipation.

\subsection*{Energy Dissipation as Consequence of Negative Feedback}
Define the discrete kinetic energy $E(t) = \frac{\Delta x}{2} \|\mathbf{u}(t)\|^2$. Its time derivative is:
\[
\frac{dE}{dt} = \Delta x \, \mathbf{u}^T \frac{d\mathbf{u}}{dt} = \Delta x \left[ \mathbf{u}^T \mathbf{f}(\mathbf{u}) + \nu \mathbf{u}^T \mathbf{A} \mathbf{u} \right]
\]
The convective term conserves energy ($\mathbf{u}^T \mathbf{f}(\mathbf{u}) = 0$ for periodic boundaries), while the viscous term gives:
\[
\mathbf{u}^T \mathbf{A} \mathbf{u} = -\frac{1}{(\Delta x)^2} \sum_{j=0}^{N-1} (u_{j+1} - u_j)^2 \leq 0
\]
Thus,
\[
\frac{dE}{dt} = -\frac{\nu}{\Delta x} \sum_{j=0}^{N-1} (u_{j+1} - u_j)^2 \leq 0
\]
The viscous term is the sole energy sink, and the negativity of $\mathbf{u}^T \mathbf{A} \mathbf{u}$ directly follows from the negative definiteness of $\mathbf{A}$ on the space of non-constant modes.

\subsection*{Control-Theoretic Interpretation}
From a control perspective, the discretized viscous term implements \textbf{\textit{distributed state feedback}}:
\[
\text{Viscous term} = \nu \mathbf{A} \mathbf{u} = \underbrace{-\frac{2\nu}{(\Delta x)^2} \mathbf{I}}_{\text{local damping}} \mathbf{u} + \underbrace{\frac{\nu}{(\Delta x)^2} \mathbf{B}}_{\text{neighbor coupling}} \mathbf{u}
\]
where $\mathbf{B}$ has zeros on the diagonal and ones on the sub/super-diagonals (with periodic wrapping). The first term provides direct negative feedback proportional to each grid point's own velocity, while the second term couples neighboring points. The overall effect drives each $u_j$ toward the average of its neighbors:
\[
(\mathbf{A}\mathbf{u})_j \propto \left( \frac{u_{j-1} + u_{j+1}}{2} - u_j \right)
\]
This is a distributed control law that minimizes velocity differences between adjacent points.

\subsection*{Stabilizing Role Against Nonlinear Instability}
Even when the nonlinear convective term $\mathbf{f}(\mathbf{u})$ may induce instability (e.g., by steepening gradients toward shock formation), the viscous term's negative feedback provides stabilization. The linearized system around any state $\mathbf{u}^*$ is:
\[
\frac{d\delta\mathbf{u}}{dt} = \mathbf{J}_{\mathbf{f}}(\mathbf{u}^*) \delta\mathbf{u} + \nu \mathbf{A} \delta\mathbf{u}
\]
where $\mathbf{J}_{\mathbf{f}}$ is the Jacobian of $\mathbf{f}$. While $\mathbf{J}_{\mathbf{f}}$ may have eigenvalues with positive real parts (indicating local instability), the addition of $\nu \mathbf{A}$---with its strictly negative eigenvalues for non-constant modes---can shift the overall system eigenvalues into the left half-plane, provided $\nu$ is sufficiently large relative to the convective time scale.



\subsection*{Conclusion}
The spatial discretization of the viscous term in the Burgers equation reveals its fundamental nature as a \textbf{\textit{linear negative feedback operator}}. The matrix $\mathbf{A}$ is negative semi-definite, with eigenvalues corresponding to modal damping rates. 
This feedback structure monotonically dissipates kinetic energy, drives the velocity field toward spatial uniformity, regularizes potential singularities arising from nonlinear convection, and inherently provides numerical stability to the discrete system.
Thus, viscosity in fluid systems---whether physical or numerical---acts as a distributed controller that enforces stability by penalizing spatial velocity variations, embodying the principle that \textbf{\textit{negative feedback creates order from chaos}}.

\section{Experiment Details and Additional Results}\label{appSec:exp-de}
This section provides a detailed overview of the parameters used in our experiments to ensure reproducibility.
All experiments were run on compute nodes, each node featuring eight A100 GPUs.

\subsection{Toy Example}\label{app:exp_toy}

\subsubsection{Motivation \& Broader Context}
The core challenge addressed in this experiment---identifying and characterizing the latent structure of multi-modal distributions from samples---is a cornerstone of unsupervised learning with profound implications across disciplines. Our goal is to develop a method that not only detects distinct modes but also robustly discriminates them from minor variations, akin to learning a parsimonious deterministic model from stochastic demonstrations. This capability is critical in diverse scenarios where the observed data is a mixture generated by several latent sources or decisions:

\begin{itemize}
    \item \textbf{Geology \& Resource Exploration}: 
    In mineral prospecting, the observed data are the spatial locations of ore fragments (e.g., gold grains) found in sediment samples. The underlying generative process is a mixture: each active deposit acts as a latent center, emitting fragments with a dispersion pattern that can be modeled by a spatial distribution (e.g., a 2D Gaussian centered at the deposit). The collected sample thus forms a \emph{multi-modal spatial point pattern}, where each mode's center indicates a potential deposit location, and the spread indicates erosion and transport effects. The key inference tasks are to determine \emph{how many deposits exist} and \emph{where they are located}, directly mirroring the problem of estimating the number and parameters of components in a Gaussian Mixture Model (GMM).

    \item \textbf{Ecology \& Wildlife Conservation}: 
    Data from camera traps or GPS collars provide sporadic location points of individuals from one or more animal populations. The distribution of these points in a landscape is intrinsically multi-modal. Each population has a preferred home range or core habitat (a latent center), resulting in a higher density of observations around it, forming one mode. Multiple populations, possibly of the same species with separate territories, give rise to multiple modes. The challenge is to deconvolve this mixture to estimate the number of distinct groups and the central point of each group's range, separating true population centers from random individual movement or outlier sightings.

    \item \textbf{Climate Science \& Atmospheric Regimes}: 
    Daily weather maps (e.g., sea-level pressure fields) represent points in a high-dimensional state space. Over time, the system does not visit all states uniformly but tends to linger around a few preferred, recurrent patterns known as \emph{weather regimes} or \emph{teleconnection patterns} (e.g., the North Atlantic Oscillation). Each regime acts as an attractor, making states near it more probable and forming a density cluster. The long-term climate record thus represents a \emph{mixture of these regimes}. Identifying these regimes involves clustering in state space to find the number of recurrent patterns (modes) and their characteristic spatial structures (centroids). This is crucial for understanding low-frequency climate variability and improving sub-seasonal forecasting.

    \item \textbf{Imitation Learning \& Robotic Decision-Making}: 
Consider an autonomous vehicle learning obstacle avoidance from human driving data. When an obstacle appears ahead, different drivers may choose entirely different strategies — such as turning left (Mode A), turning right (Mode B), or braking to a stop (Mode C) — resulting in a multimodal distribution in the action space. A simple behavioral cloning model trained to minimize average error would converge toward the average of these modes, likely producing an unsafe, indecisive response, such as hesitant steering or delayed braking. Additionally, within each strategic mode — for example, all left-turn trajectories — there exists natural variation in execution, such as slight differences in steering angle or timing. A robust method must therefore distinguish between meaningful strategic modes and treat execution-level variations as noise to be filtered out rather than as separate strategies, capturing the essential challenge of extracting clear multimodal policy structure from noisy demonstration data.
\end{itemize}

In all these cases, the \emph{number of contributing sources or high-level decisions is unknown a priori}. A successful method must perform dual inference: estimating this number and robustly identifying the parameters of each major component while being robust to within-mode noise. Our experimental framework is designed to rigorously evaluate this critical capacity for \emph{recovering true latent structure from mixture samples}.

\subsubsection{More Discussion on Reconstruction Error Optimization in Precision-Demanding Tasks}\label{appSec:preciseTask}

Generative models utilize probabilistic modeling. The benefit of this approach is that once a distribution is learned, more new samples can be sampled, thus accomplishing the task of generation. 
However, in scenarios such as robot manipulation or temporal prediction, our primary objective is to model and understand the system, rather than approximating the underlying distribution and generating highly diverse and stylistically varied samples as is common in computer vision (CV).
For instance, in imitation learning for robotic manipulation, expert human demonstration data often contains inherent jitter or other artifacts. Our objective is for the model to learn a smooth and robust policy, instead of modeling this underlying noise distribution.
A mainstream framework in robotic manipulation is to feed observed images as input and predict the robot's next action(s).
This resembles a traditional supervised learning task (instead of probabilistic modeling), where we learn a mapping from inputs to outputs from the data.
What differs is that the data distribution here often exhibits multi-modal properties.
Taking the robot obstacle avoidance problem as an example: to avoid an obstacle ahead, the robot can detour by moving either left or right. These two options create a bimodal distribution in the data. If we simply use an Mean Squared Error (MSE) loss between the model's output and the collected sample data, the model may only learn an average behavior — ultimately causing it to move straight forward and collide with the obstacle.
Therefore, Generative models such as diffusion \citep{chi2023diffusion} are used to capture this multi-modal nature in the data.

In summary, while traditional frequentist supervised learning enables us to discard the noise term and achieve high precision, probabilistic models allow us to capture more complex data distributions. 
But within each mode of the data distribution, we do not want to compromise on precision.
In fact, the approach we adopt to conditional flow matching—via joint learning or optimal transport—achieves semantic segmentation of the noise space $x_0$ \citep{pooladian2023multisample,tong2024improving}. 
Then our MLE fine-tuning based on the reconstruction results of each mode effectively compensates for the precision requirements in probabilistic models.
The demand for precision necessitates the consideration of reconstruction error. This may also partially  explain why VAE-based policy ACT \citep{zhao2023learning} has achieved significant success in robot fine manipulation tasks, despite being less capable than diffusion or flow-based models in capturing action diversity.

\subsubsection{Experimental Details}

\paragraph{Target Distribution Configuration.}
The target distribution is a 2D mixture Gaussian distribution with five centers at coordinates $[-0.8, -0.5]$, $[-0.1, -0.3]$, $[0.6, -0.4]$, $[-0.4, 0.8]$, and $[0.5, 0.7]$. The mixture weights are non-uniform: $[0.3, 0.2, 0.2, 0.1, 0.2]$. Each component has a standard deviation $\sigma = 0.005$ with diagonal covariance matrix $\sigma^2I = 0.0001 \times I$.

\paragraph{Model Architecture.}
We employ a FM model based on Optimal Transport Conditional Flow Matching (OT-CFM). The network is a 3-layer fully connected neural network with 128 hidden dimensions and ReLU activation functions. The model takes 2D spatial coordinates concatenated with time $t$ as input and outputs a 2D velocity field vector.

\paragraph{Pretraining Hyperparameters.}
The model is pretrained for 2000 epochs with a batch size of 512. We use the Adam optimizer with a learning rate of 0.001 and cosine annealing learning rate scheduler. The loss function is the flow matching loss with OT pairing mode and $\sigma = 0.0$ for deterministic flow matching. 

\paragraph{MLE Fine-tuning Hyperparameters.}
For MLE fine-tuning, we train for 1000 epochs with batch size 512. The learning rate is 0.001 using Adam optimizer with weight decay $1\times10^{-7}$ and cosine annealing scheduler. The loss function is MSE loss with 10 sampling steps and sorted pairs pairing mode. Fine-tuning starts from the pretrained checkpoint at epoch 1000.

\subsubsection{Additional Results}
\begin{figure}[t]
	\centering
	\includegraphics[width=1\textwidth]{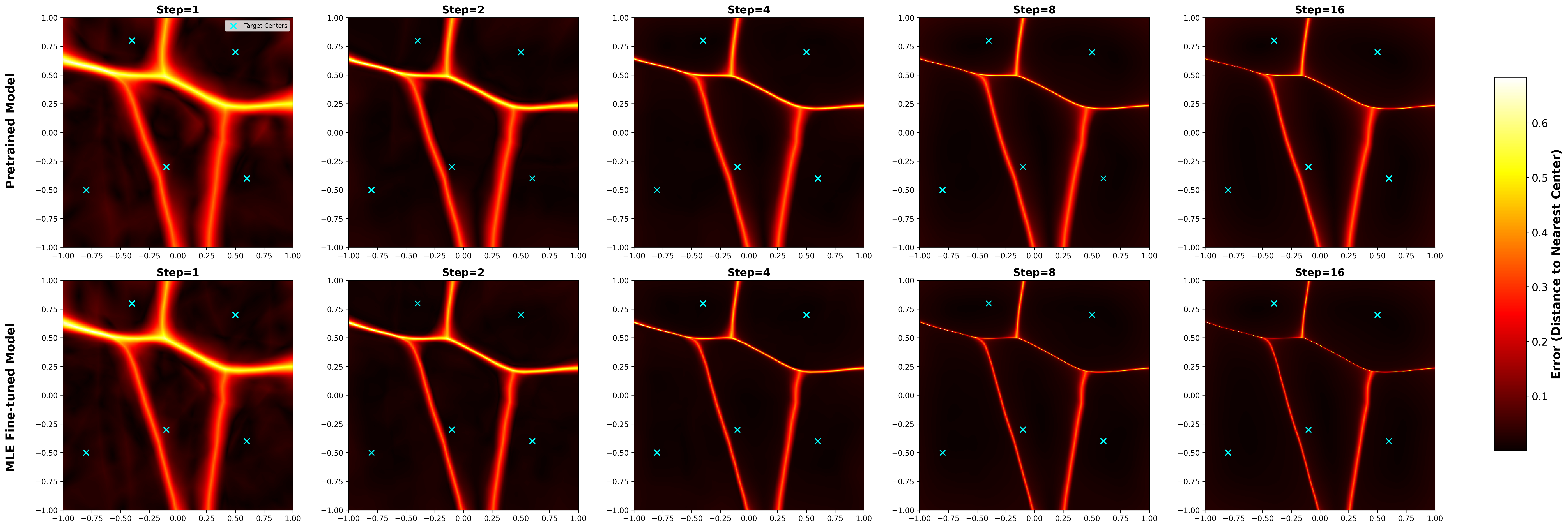}
	\caption{Error visualization across the latent space: a comparison of pre-trained and fine-tuned models at different numerical precisions (Extended Version with Additional Details)}
	\label{fig:numeric_error_comparison_heatmaps_detailed}
\end{figure}

\begin{figure}[t]
	\centering
	\includegraphics[width=0.45\textwidth]{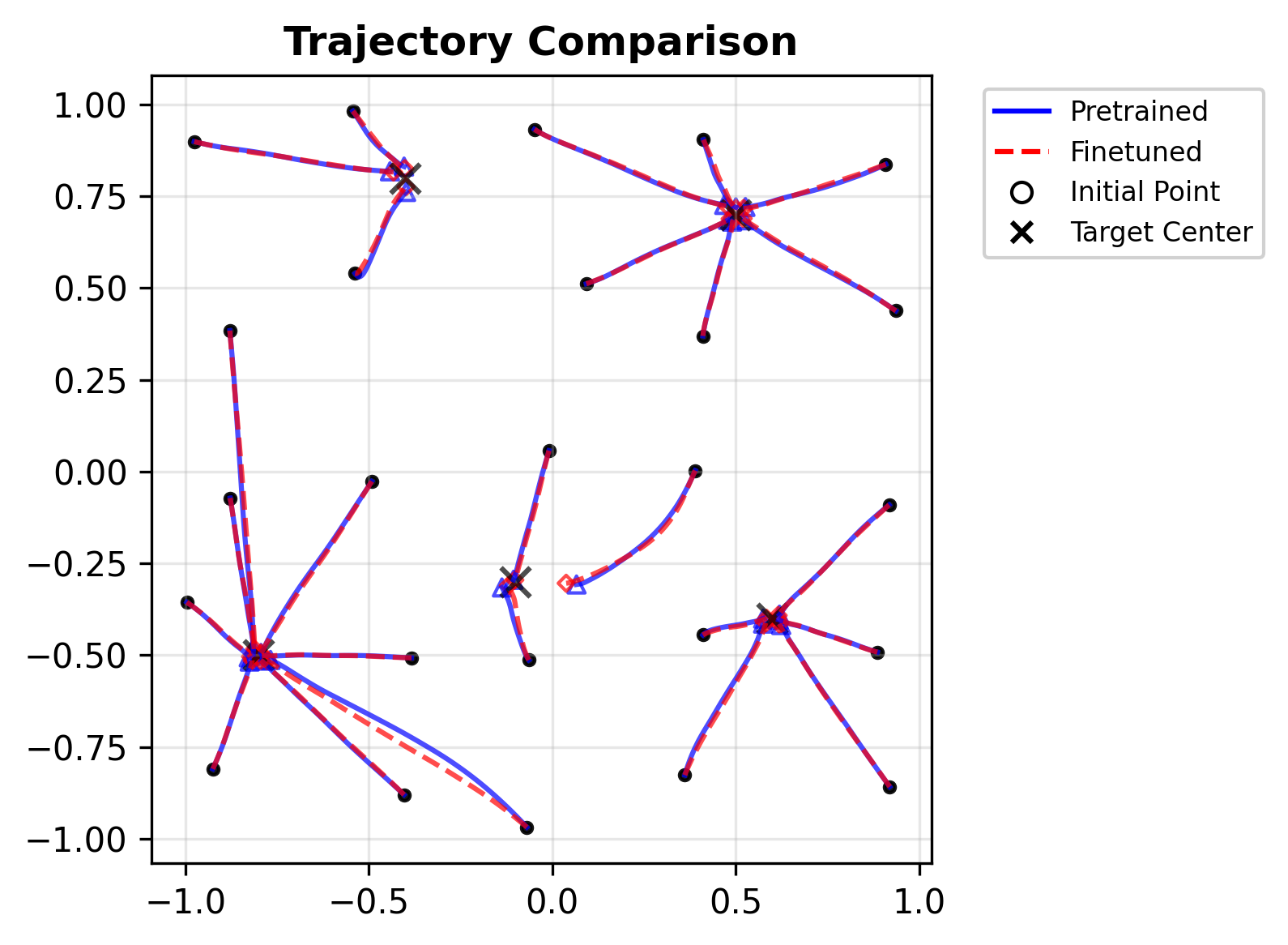}
	\caption{trajectories}
	\label{fig:trajectories-app}
\end{figure}

Figure \ref{fig:numeric_error_comparison_heatmaps_detailed} illustrates the error distributions of the Euler discretization method under different step sizes (1, 2, 4, 8, 16).
MLE Fintuning consistently achieved better inference results across different scales.
Figure \ref{fig:trajectories-app} draws the inference trajectories of the pre-trained and fine-tuned models starting from some randomly selected initial points. 
It can be observed that after fine-tuning, the trajectory endpoints generally move closer to the center. Additionally, fine-tuning does not increase the curvature of the trajectories; in fact, some originally curved trajectories even become straighter after fine-tuning. 

\subsection{Robotic Manipulation}

\subsubsection{Experimental Details}

We employ two types of network architectures to learn the vector field:
a Conditional UNet-1D and a Transformer-based model. The UNet model was designed with an input dimension matching the action space of the environment and an output dimension corresponding to the action dimensions. It incorporated global conditioning based on visual feature encodings, a hidden size of 256, 4 UNet blocks with 2 layers each, and attention mechanisms to enhance representational capacity. The Transformer model used a hidden dimension of 512, 6 transformer layers, 8 attention heads, and a dropout rate of 0.1 to prevent overfitting. Both models utilized a ResNet-18 vision encoder for extracting image features, with GroupNorm substituted for BatchNorm to improve training stability. In the residual fine-tuning part, the normalization terms use the same architecture as in the pre-training phase, but with fewer hidden layers and parameters, and with the state input set to zero for decoupling. In the ablation study, the GAV term is replaced with a single-hidden-layer MLP comprising 128 hidden nodes. For sampling, DDPM uses 100 steps, while all other methods (DPM, DDIM, FM, etc.) employ 16 steps. Additional details can be found in the source code.

We used a batch size of 64. In the pre-training stage the Adam optimizer was used with a learning rate of 1.0e-4 and weight decay of 1.0e-6, accompanied by a linear warm-up phase spanning 500 steps. 
While in MLE fine-tining stage we use a learning rate of 5.0e-6. 
In the design for inducing contraction properties we choose $\lambda_{\omega}=0.05$ and $\epsilon_A=10^{-6}$.
All models were trained on 4 A100 GPUs using a Distributed Data Parallel strategy with 32-bit precision. An Exponential Moving Average was applied with a decay rate of 0.75, updated every 50 epochs. 

Environment-specific parameters were carefully tailored to each experimental setting. For the kitchen environment, the observation horizon was set to 2, the prediction horizon to 24, and the action horizon to 8, with an action dimension of 9 and a visual feature dimension of 512. Each episode was allowed a maximum of 280 steps. In the Push-T environment, the observation horizon was 1, prediction horizon 16, and action horizon 8, with an action dimension of 2 and visual feature dimension of 514. The maximum steps per episode were 300. For the Mimic environment, the observation horizon was 1, prediction horizon 32, and action horizon 16, with action dimension 7 and visual feature dimension 512. Episodes were run for up to 400 steps.


Table~\ref{tab:hyperparameters} shows the hyperparameters used in FM and diffusion policy. Table~\ref{tab:taskparameters} shows the task summary.
Our evaluations were all conducted within the MuJoCo or Gym simulation environment. 
Specifically, the Push-T simulation environment is developed using Gym, whereas the Franka Kitchen and Robomimic environments utilize MuJoCo for simulation.

\begin{table*}[t]
	\centering
	\resizebox{0.9\textwidth}{!}{
		\begin{tabular}{lrrrrrrrrrr}
			\toprule[1.5pt]
			\textbf{H-Param}                      & \textbf{Ta}      & \textbf{Tp}           & \textbf{ObsRes}       & \textbf{F-Net}                            & \textbf{F-Par}
			& \textbf{V-Enc}                      & \textbf{V-Par}  & \textbf{P-Lr}           & \textbf{F-Lr}         
			\\ \midrule			
			Franka Kitchen                        & 8                & 16                    & 1x60                   & ConditionalUnet1D & 66                    &    N/A                   
			& N/A               & 1e-4                                & 5e-6                  
			\\
            Push-T                                & 8                & 16                    & 1x96x96                 & ConditionalUnet1D & 80                    & ResNet-18             & 11                & 1e-4                                & 5e-6                  
			\\
			Robomimic                             & 8                & 16                    & 1x50                   & ConditionalUnet1D & 66                    & N/A                   & N/A               & 1e-4                                & 5e-6                        
			\\\toprule[1.5pt]
	\end{tabular}}
	\caption{Hyperparameters for FM and diffusion policy. Ta: action horizon. Tp: action prediction horizon. ObsRes: environment observation resolution. D-Net: diffusion/flow matching network. D-Par: diffusion/flow matching network number of parameters in millions. V-Enc: vision encoder. V-Par: vision encoder number of parameters in millions. P-Lr: learning rate in pretraining. F-Lr: learning rate in MLE fine-tuning.} 
	\label{tab:hyperparameters}
\end{table*}

\begin{table}[!t]
	\centering
	\resizebox{0.5\columnwidth}{!}{
	\begin{tabular}{lrrrrrr}
			\toprule[1.5pt]
			\textbf{Tasks}   & \textbf{Rob} & \textbf{Obj} & \textbf{ActD} & \textbf{PH}  & \textbf{Steps} & \textbf{Img}    
			\\ \midrule
			Push-T           & 1            & 1            & 2             & 200          & 300            & \cmark               \\
			Franka Kitchen   & 1            & 7            & 9             & 566          & 280            & \xmark            \\
			Robomimic        & 2            & 3            & 20            & 200          & 700            & \xmark        
			\\\toprule[1.5pt]
	\end{tabular}}
	\caption{Tasks Summary. Rob: number of robots. Obj: number of objects. ActD: action dimension. PH: proficient-human demonstration. Steps: max number of rollout steps. Franka Kitchen and Robomimic involve 6D robot and gripper actions in the joint space. Push-T focuses on robot end-effector trajectories.}
	\label{tab:taskparameters}
\end{table}

\begin{figure}[h]
	\centering
	\begin{adjustbox}{scale=1}
		\begin{subfigure}[b]{0.2\linewidth} 
			\centering
			\includegraphics[width=\linewidth]{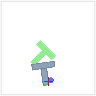}
			\caption{Push-T}
			\label{fig:pushT-env}
		\end{subfigure}		
		\begin{subfigure}[b]{0.35\linewidth} 
			\centering
			\includegraphics[width=\linewidth]{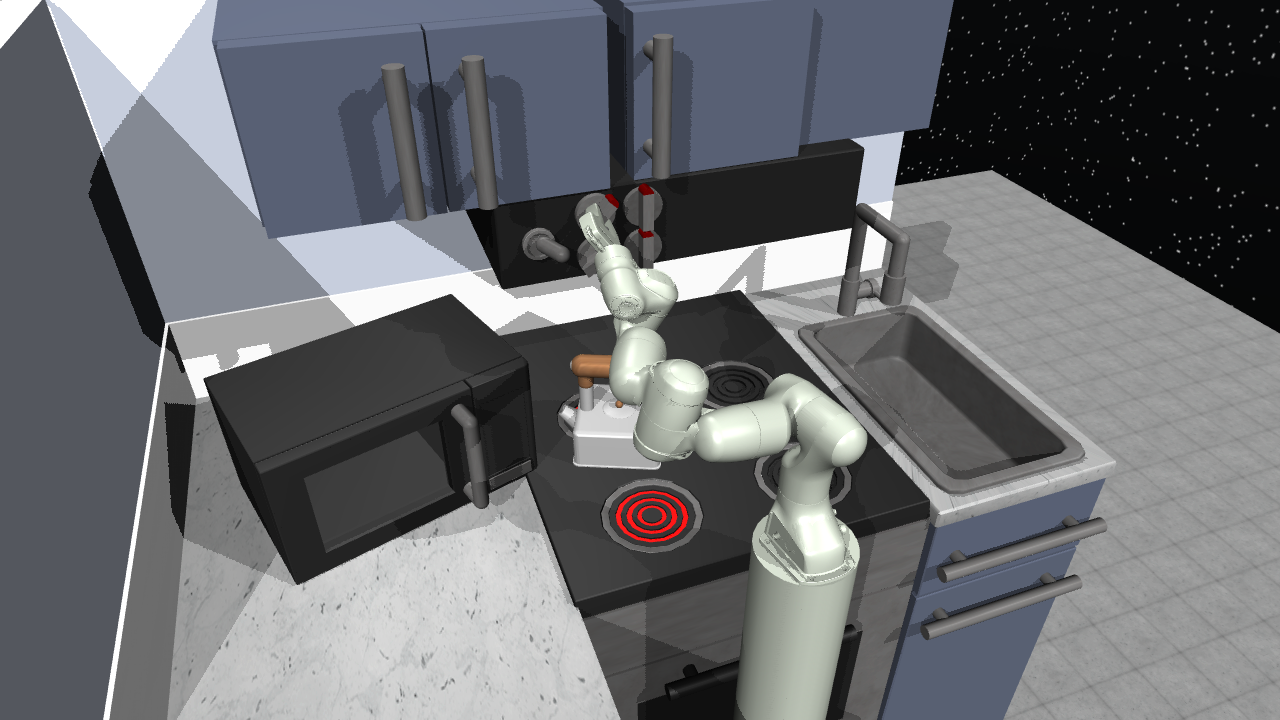}
			\caption{Franka Kitchen}
			\label{fig:kitchen-env}
		\end{subfigure}		
		\begin{subfigure}[b]{0.197\linewidth}
			\centering
			\includegraphics[width=\linewidth]{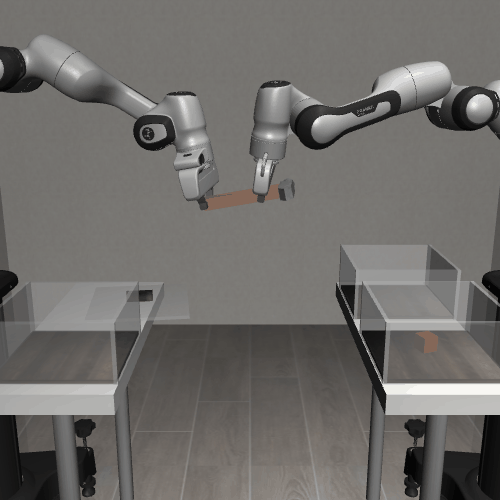}
			\caption{Robomimic}
			\label{fig:mimic-env}
		\end{subfigure}
	\end{adjustbox}
	\caption{Different Simulation Environments in MuJoCo and Gym.
	}
	\label{fig:mujoco}
\end{figure}

\subsubsection{Additional Results}
\begin{figure*}[t]
	\centering
	\includegraphics[width=1\textwidth]{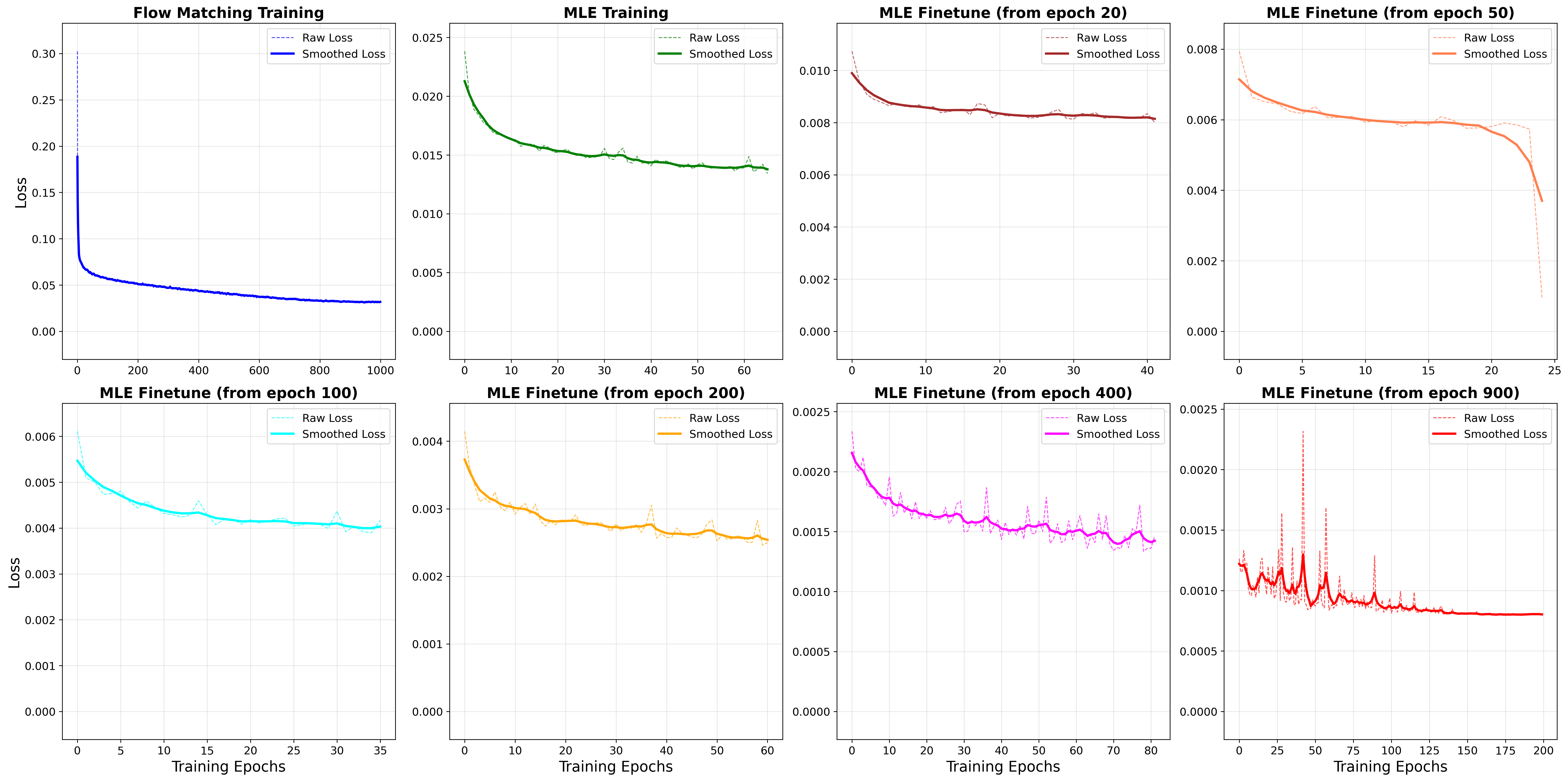}
	\caption{Training Loss curves of MLE Fine-tuned Models at Different Stages in the Franka Kitchen Environment.}
	\label{fig:kitchen_curve_loss}
\end{figure*}

Figure~\ref{fig:kitchen_curve_loss} presents the loss curves corresponding to the results shown in Figure~\ref{fig:kitchen_sub}. 
It can be observed that, compared to ERA5 (Figure~\ref{fig:era5_sub}), the MLE fine-tuning training error curve for the robot policy is smoother, indicating a relatively lower training difficulty. This is precisely because the generated variable dimension of the robot policy is relatively low (11-dimensional in the Franka Kitchen environment), which reduces the difficulty of ODE simulation. However, compared to FM pre-training, the training difficulty is still somewhat higher.
In the fourth subplot (where fine-tuning starts at 50 epochs), after going through a phase of optimization plateau, the training error declines rapidly toward the end, corresponding to the relatively noticeable rise in the final segment of the success rate curve shown in Figure~\ref{fig:kitchen_sub}.

\subsection{Meteorological Data}\label{app:era5}
\subsubsection{Experimental Details}
The experimental evaluation employs the Flow Neural Process (FNP) model on ERA5 meteorological reanalysis data. 

ERA5 data for the US region are normalized and preprocessed with circular handling, providing 950--1000 randomly sampled context points per instance. The FNP model architecture incorporates a continuous normalizing flow with Transformer-based attention mechanisms and high-frequency positional encoding.

Base training runs for 200 epochs with a batch size of 100, learning rate $5\times10^{-4}$, and cosine annealing. Validation occurs every 20 epochs, and checkpoints are saved accordingly. The MLE fine-tuning phase loads the best base checkpoint and continues for 100 epochs with a reduced learning rate $1\times10^{-5}$, batch size 32, weight decay $1\times10^{-7}$, and 16 flow-matching sampling steps. Fine-tuning employs distributed training across five GPUs with linear warmup and constant scheduling.
During residual fine-tuning, we retain the pre-training normalization architecture but with a reduced scale (fewer layers and parameters) and zero-initialized state input to ensure decoupling. In contrast, the ablation study replaces the GAV module with a simpler, single-hidden-layer MLP (128 nodes). For complete implementation specifics, consult the source code.
Evaluation uses 50 Monte Carlo samples to compute log-likelihood, mean squared error, and uncertainty calibration.

Baseline Models include Conditional Neural Process (CNP) \cite{garnelo2018conditional}, Neural Process (NP) \cite{garnelo2018neural}, attention mechanisms Neural Process (ANP) \cite{kimattentive} and Transformer Neural Process (TNP) \cite{nguyen2022transformer}. The implementation employs \citet{nguyen2022transformer,lee2020bootstrapping}.
\subsubsection{Additional Results}
\begin{figure}[htbp]
    \centering
    \subfloat{\includegraphics[width=0.2\textwidth]{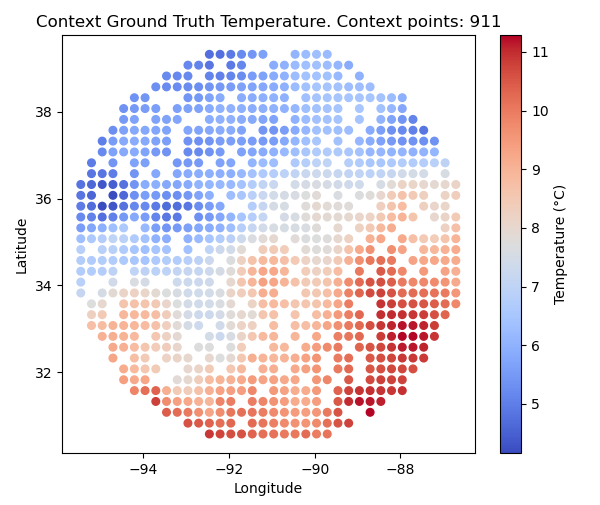}}
    \hfill
    \subfloat{\includegraphics[width=0.2\textwidth]{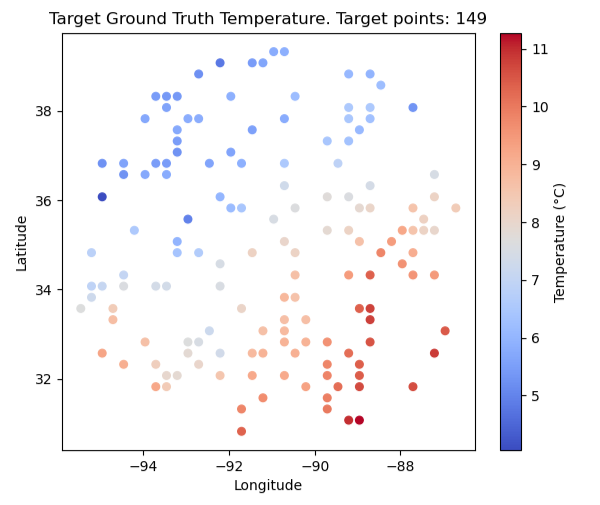}}
    \hfill
    \subfloat{\includegraphics[width=0.2\textwidth]{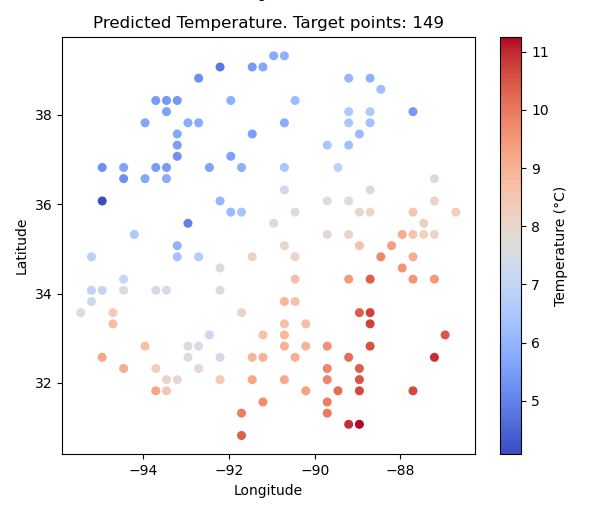}}
    \hfill
    \subfloat{\includegraphics[width=0.2\textwidth]{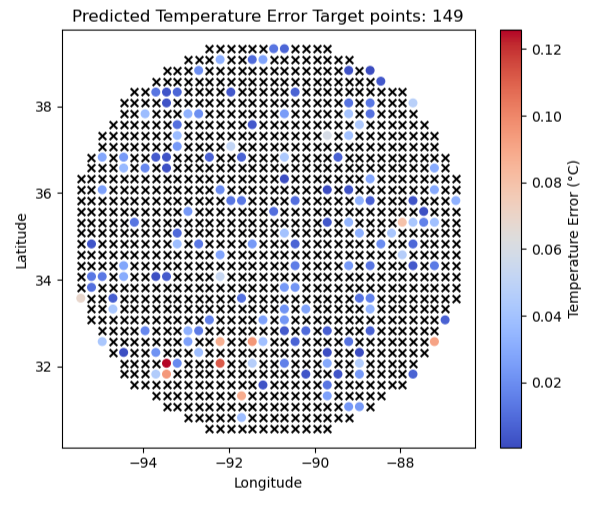}}
    \hfill
    \subfloat{\includegraphics[width=0.18\textwidth]{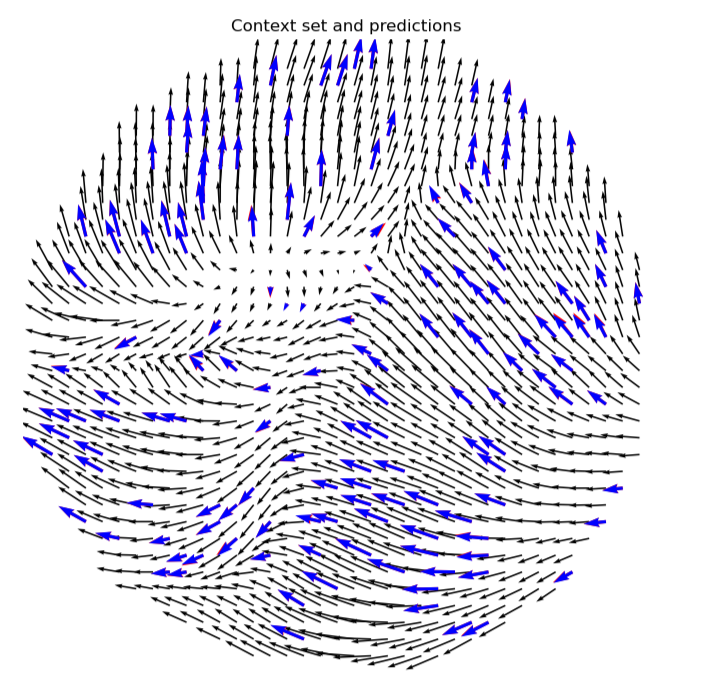}}
    \caption{Visualization of predicted results in ERA5.}
    \label{fig:era5_pred_full}
\end{figure}

\begin{figure*}[t]
	\centering
	\includegraphics[width=1\textwidth]{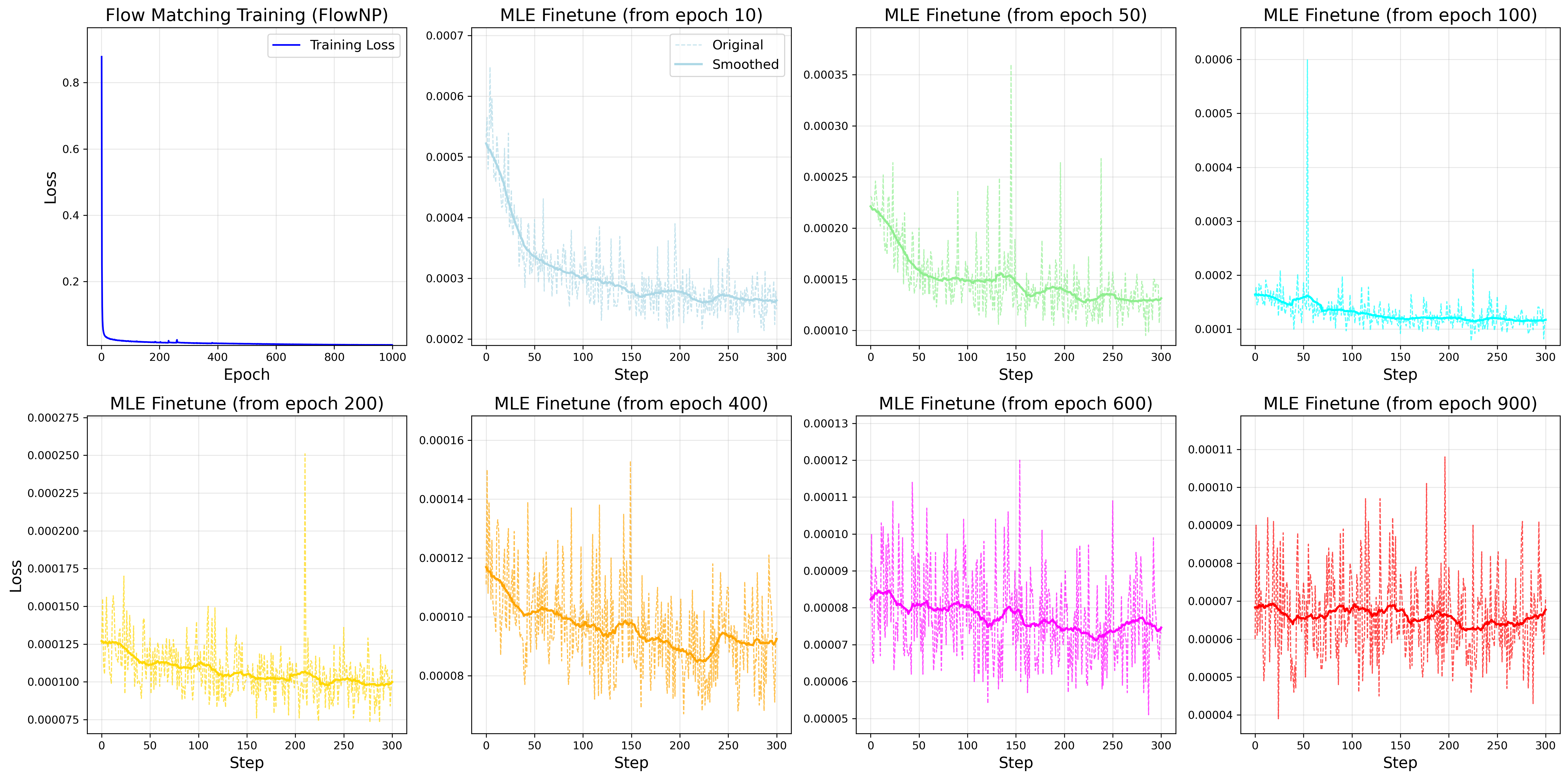}
	\caption{Training Loss curves of MLE Fine-tuned Models at Different Stages in the ERA5 dataset.}
	\label{fig:era5_curve_loss}
\end{figure*}

Figure \ref{fig:era5_pred_full} presents the visualization results of an experiment. The first four subplots depict temperature, while the last one illustrates wind speed.
Figure \ref{fig:era5_curve_loss} presents the loss curves corresponding to the results shown in Figure \ref{fig:era5_sub}.

\subsection{Image Generation}
We perform an experiment on unconditional CIFAR-10 generation from a Gaussian source to examine how OT-CFM performs in the high-dimensional image setting. We use a similar setup to that of \cite{lipman_flow_2022} and \cite{tong2024improving}, including the time-dependent U-Net architecture from \citet{nichol2021improved} that is commonly used in diffusion models. We choose channels $=128$, depth $=2$, channels multiple $=[1,2,2,2]$, heads $=4$, heads channels $=64$, attention resolution $=16$, dropout $=0.1$, batch size per gpu $= 128$, gpus $=4$, epochs $=2000$.
We use a constant learning rate, set to $2\times10^{-4}$ in the  pre-training stage and $5\times10^{-6}$ in the fine-tuning stage. To prevent training instabilities and variance, we clip the gradient norm to 1 and rely on exponential moving average with a decay of 0.99. 

The baseline models are chosen as DDPM \citep{ho2020denoising}, FM
with Variance Exploding (VE) path and Optimal Transport (OT) path \citep{lipman_flow_2022}, and Stochastic Interpolants (SI) \citep{albergo2023stochastic}.
We train our fine-tuned FM and report the Fr\'echet inception distance (FID) in \autoref{tab:cifar}. 

From \autoref{tab:cifar}, we can see that fine-tuning FM can improve the performance on FID with almost the same number of function evaluations (NFE).
These demonstrate the significance of fine-tuning FM.

\begin{table}[t!]
	\centering
	\raisebox{-0.5\baselineskip}{
		\begin{minipage}{0.44\textwidth} 
			\caption{FID score and number of function evaluations (NFE) for different ODE solvers: fixed-step Euler integration with 100 and 1000 steps and adaptive integration \citep[\textsc{dopri5}]{hairer1993solving}. The adaptive solver is significantly better than the Euler solver in fewer steps. {First results are from \cite{lipman_flow_2022} and the next three from \cite{tong2024improving}}. The last rows report the results of our fine-tuned FM. }
			\label{tab:cifar}
	\end{minipage}}\hfill
	\begin{minipage}{0.54\textwidth}\hfill  
		\begin{tabular}{@{}lcccc}
			\toprule[1.5pt]
			NFE / sample $\rightarrow$  & \multicolumn1c{100} & \multicolumn1c{1000} & \multicolumn{2}{c}{Adaptive}\\\cmidrule(lr){2-2}\cmidrule(lr){3-3}\cmidrule(lr){4-5}
			Algorithm $\downarrow$  & \multicolumn1c{FID} & \multicolumn1c{FID} & \multicolumn1c{FID} & \multicolumn1c{NFE} \\
			\midrule
			{DDPM} & & & {7.48} & {274} \\
			VP-FM  & {7.772} & {4.048} &  {4.335}          & {525.92}\\
			OT-FM  & {4.640} & {3.822} &           3.655 & 143.00  \\
			S.I. & {4.488} & {4.132} &     {4.009}       & {146.12} \\
			I-CFM & {4.461} & 3.643 &           3.659 & 146.42 \\
			OT-CFM & \textbf{4.443} & {3.741} &  3.577 & \textbf{133.94}\\
			FT-FM (ours)  & 4.451 & \textbf{3.620} &           \textbf{3.496} & 146.42 \\
			\bottomrule[1.5pt]
		\end{tabular}
	\end{minipage}
\end{table}

For sampling, we use Euler integration using the torchdyn package and \textsc{dopri5} from the torchdiffeq package. Since the \textsc{dopri5} solver is an adaptive step size solver, it uses a different number of steps for each integration. We use a batch size of 1024 and average the number of function evaluations (NFE) over batches.

\begin{figure}[t]
	\centering
	\begin{subfigure}{\linewidth}
		\centering
		\includegraphics[width=\linewidth]{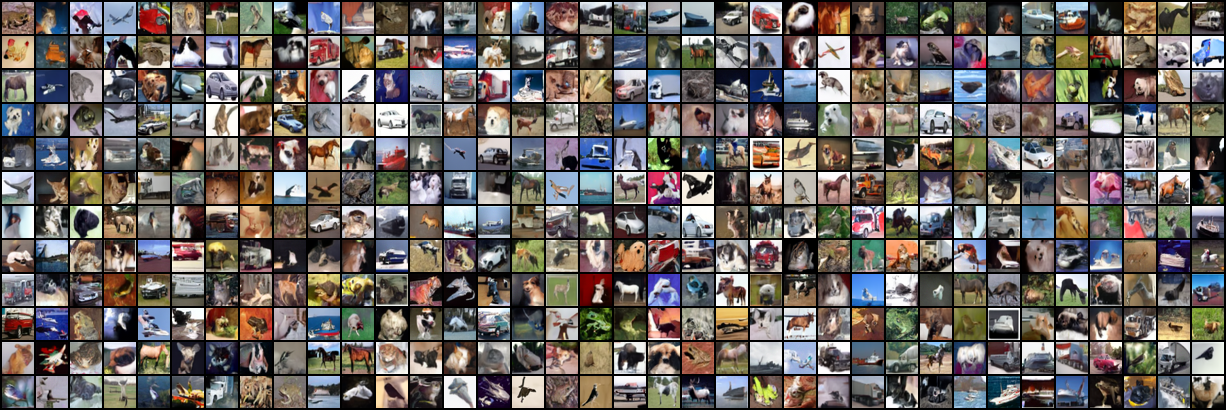}
		\caption{Images generated by FM at $10^5$ step, FID: 6.17.}
		\label{fig:Cifar10-1}
	\end{subfigure}
	\begin{subfigure}{\linewidth}
		\centering
		\includegraphics[width=\linewidth]{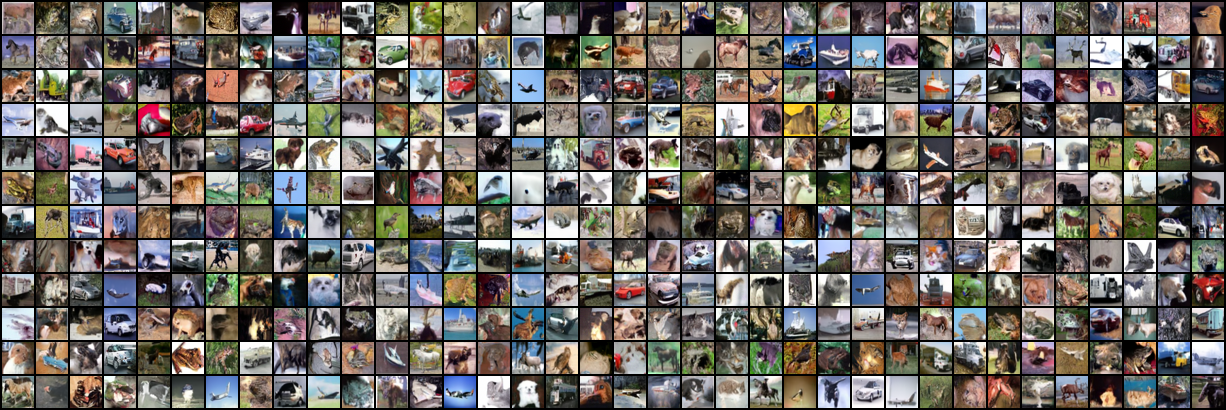}
		\caption{Images generated by FM at $3.5 \times 10^5$ step, FID: 3.57.}
		\label{fig:Cifar10-2}
	\end{subfigure}
	\\
	\begin{subfigure}{\linewidth}
		\centering
		\includegraphics[width=\linewidth]{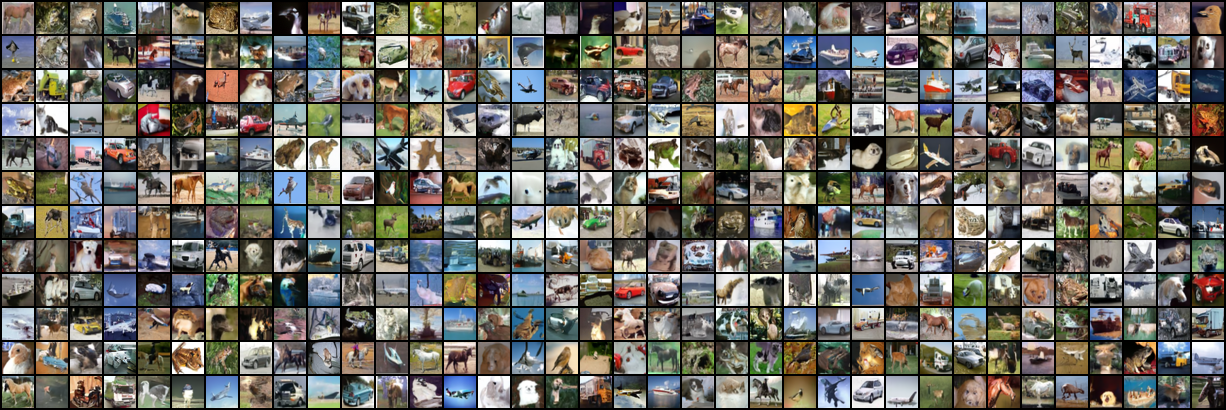}
		\caption{Images generated by our fine-tuned FM model. Fine-tuning started from the $3.5 \times 10^5$ step and continued for $1,000$ steps, FID: 3.55.}
		\label{fig:Cifar10_ft}
	\end{subfigure}
	\caption{Visualization Results of generated images under the same initial noise points.}
	\label{fig:Cifar10_image}
\end{figure}

The visualization results of the generated images are presented in Figure \ref{fig:Cifar10_image}.
We can see that fine-tuning induces minimal alteration to the latent space semantics. Images generated from identical noise points retain consistent content, with a slight improvement in reconstruction fidelity.
From Table \ref{tab:cifar}, we can see that fine-tuning FM can  	slightly improve the performance on FID with almost the same NFE.
The effectiveness on image data, however, is less pronounced compared to the results on policy and other tasks presented previously.
This is because the generative variables for images are high-dimensional ($32\times32\times3 = 3071$ in this case), and pixels exhibit strong, complex correlations. This significantly increases the difficulty of the required MLE training.
In future work, we plan to conduct generation experiments within the compressed latent space.


\end{document}